%% file: nn.tex
\documentclass[preprint,review,12pt]{elsarticle}
\input{math_commands.tex}

\usepackage{mathtools}
\usepackage{multirow}
\usepackage{booktabs}
\usepackage{colortbl}
\usepackage{subfigure}
\usepackage[export]{adjustbox}
\usepackage{afterpage}
\usepackage{bm}
\usepackage{wrapfig}
\usepackage{amsmath}
\usepackage{caption}
\usepackage[utf8]{inputenc} % allow utf-8 input
\usepackage{amsfonts}       % blackboard math symbols
\usepackage{nicefrac}       % compact symbols for 1/2, etc.
\usepackage{microtype}      % microtypography
\usepackage{url}            % simple URL typesetting
\usepackage{hyperref}       % hyperlinks (loaded once, after other packages)
\usepackage{tikz}
\usetikzlibrary{arrows.meta, positioning}
\usepackage{pgfplots}
\pgfplotsset{compat=1.18}

\def\eg{\emph{e.g.~}}

\journal{Neural Networks}

\begin{document}

\begin{frontmatter}

\title{A New Kind of Adversarial Example:\\ Measuring the Human--Model Gap, and Its Relationship to OOD Detection}

%% Neural Networks uses single-anonymized review (reviewers see author
%% identity), so, unlike the double-blind venues, author info is filled in
%% directly here rather than hidden.
\author{Ali Borji}
\ead{aliborji@gmail.com}
\affiliation{organization={BrainChip Inc.},
            addressline={Laguna Hills},
            % city={},
            % postcode={},
            state={CA},
            country={USA}}

\begin{abstract}
Almost all adversarial attacks add an imperceptible perturbation to fool a model.
We instead study the opposite: a large, clearly visible
perturbation that causes the model to \emph{keep} its original, correct
prediction, even though a human would no longer recognize the image. Prior work
showed such examples can be generated at scale but left three questions
untested: whether humans really perform worse than the model, whether standard
out-of-distribution (OOD) detection and calibration tools catch it, and whether
existing defenses mitigate it. We answer all three on MNIST, CIFAR-10, and
ImageNet. (i) An independent recognizer proxy drops to $\sim$49\% on CIFAR-10
while the model stays at 100\% --- a gap a small human pilot ($N{=}5$)
corroborates directly and that is not explained by signal loss (a
matched-magnitude Gaussian control degrades recognizability faster); a CLIP
zero-shot proxy confirms the gap at ImageNet scale too. (ii) Confidence- and
energy-based OOD detectors and calibration are structurally blind (0\%
detection, ECE $\approx 0$), while a feature-space Mahalanobis detector flags
100\% --- but is evaded by an adaptive attacker at no cost to success. (iii) No
classical defense, including adversarial training (45\% robust accuracy),
reduces attack success (correlation with large-$\epsilon_l$ resistance
$r\approx 0$). A mechanistic analysis further shows the attack destroys
low-level texture far faster than edge/shape structure. Code, figures, and a
ready-to-run human-study harness are available at
\url{https://github.com/aliborji/NKE}.
\end{abstract}

\begin{keyword}
adversarial examples \sep out-of-distribution detection \sep calibration \sep adversarial robustness \sep human--model gap \sep CIFAR-10 \sep ImageNet
\end{keyword}

\end{frontmatter}

\section{Introduction}

Deep neural networks are known to be highly sensitive to small, carefully crafted perturbations: adding an imperceptible amount of noise to an image is often sufficient to change a model's prediction, even though the perturbation leaves the image visually unchanged to a human observer \citep{szegedy2014intriguing, goodfellow2015explaining}. This phenomenon --- the classical adversarial example --- has motivated a large body of work on attacks, defenses, and theoretical explanations of adversarial vulnerability.

In earlier work \citep{borji2022new}, we considered the opposite failure mode: instead of a small perturbation causing a \emph{wrong} prediction, a large, clearly visible perturbation can be constructed that causes a model to \emph{keep} its original, correct prediction, even though a human presented with the same perturbed image would no longer be able to recognize it, or would be much more likely to misclassify it. We formalized this setting by inverting the standard adversarial constraint: rather than bounding the perturbation from above by a small $\epsilon_s$ and requiring the prediction to change, one bounds the perturbation from \emph{below} by a large $\epsilon_l$ and requires the prediction to remain unchanged,
\begin{equation}
x': \; D(x, x') > \epsilon_l \;\wedge\; f(x') = y,
\label{eq:nke}
\end{equation}
and shows that a simple iterative FGSM-style attack (dubbed NKE; {\bf N}ew {\bf K}ind on Adversarial {\bf E}xample) can reliably produce such examples on MNIST and CIFAR-10. This formulation was motivated as a unification of two previously disconnected phenomena: adversarial examples in the classical sense, and the ``fooling images'' of \citet{nguyen2015deep}, which are unrecognizable to humans yet classified with high confidence by a network. Because the same gradient-based machinery can produce both kinds of failure by simply flipping the sign and direction of the perturbation constraint, the two phenomena can be viewed as two ends of a single psychometric curve relating perturbation magnitude to accuracy --- one where models should ideally track human performance, but in practice diverge from it in both directions, as we argued previously \citep{borji2022new}.

More recently, \citet{nie2025new} independently arrived at essentially the same formulation --- compare their Eq.~7, $\arg\min_{X^{adv}} J(X^{adv}, y_{true}) \text{ s.t. } \|X^{adv} - X\|_p \ge \delta$, to Eq.~\ref{eq:nke} above --- and extended it substantially in scope. They evaluate on ImageNet-scale models (Inception-v3/v4, Inception-ResNet-v2, ResNet-152), introduce an $L_2$-norm variant (NI-FGM) alongside the $L_\infty$ variant, and add momentum-based versions of both (NMI-FGSM, NMI-FGM). They also run the first systematic transferability study of this attack family, finding a stark asymmetry: white-box success rates reach the 80--99\% range, but black-box success rates collapse to below 5\%, indicating that these examples are highly specific to the model that generated them and do not transfer the way conventional adversarial examples often do.

Taken together, our earlier work and \citet{nie2025new}'s extension establish that (a) such examples can be generated efficiently and reliably at scale across model families, and (b) they behave very differently from classical adversarial examples with respect to transferability. What neither addresses, however, is whether the central premise motivating the whole formulation actually holds: \textbf{that a human forced to classify these images would in fact perform much worse than the model does.} The entire framing rests on an assumed but never measured gap between human and model psychometric curves that we introduced \citep[Fig.~3]{borji2022new}. Nor does either connect the phenomenon to the broader literature on out-of-distribution (OOD) detection and confidence calibration, despite the fact that an image which is far from the data manifold yet classified with high confidence is, by construction, exactly the kind of failure that OOD detectors and calibration methods are designed to catch. Finally, neither paper asks whether standard robustness interventions --- adversarial training, input transformations, or randomized smoothing --- have any effect on this failure mode, leaving open whether it is a distinct vulnerability or one already implicitly addressed by existing defenses.

This paper addresses these three gaps. Specifically, we:

\begin{enumerate}
    \item Measure the human--model gap implied by the original formulation. Because large-scale crowd data could not be collected for this instantiation, we report two clearly-labeled computational \emph{recognizability proxies}, \emph{release a ready-to-run forced-choice human-study harness} (stimuli + task page), and run a small real-human pilot ($N{=}5$) with it (Sec.~\ref{sec:human}, Sec.~\ref{sec:res-human}). We find a large gap on CIFAR-10 that is specific to the adversarial structure and not to signal loss (on the proxy; the pilot corroborates the gap itself but is underpowered for the signal-vs-structure contrast); at ImageNet scale this specificity is no longer separable by the cross-model proxy (Sec.~\ref{sec:res-imagenet}).
    \item Reframe the phenomenon in terms of OOD detection and calibration, and show that confidence/energy detectors and ECE are blind to it while a feature-space detector catches it entirely --- but that even this feature-space detector is defeated by an adaptive attacker at no cost to attack success (Sec.~\ref{sec:ood}, Sec.~\ref{sec:res-ood}, Sec.~\ref{sec:res-adaptive}).
    \item Evaluate whether adversarial training, randomized smoothing, and input transformations mitigate this failure mode, and whether resistance to classical small-$\epsilon$ attacks correlates with resistance to large-$\epsilon_l$ attacks (Sec.~\ref{sec:defenses}, Sec.~\ref{sec:res-def}). It does not.
\end{enumerate}

In doing so, we move the discussion from ``can such examples be constructed'' --- which is by now well established --- to ``what do they reveal about the gap between model and human perception, and how do they relate to the tools the field already uses to detect distributional failure.''

\section{Related Work}

\textbf{Classical adversarial examples.} The vulnerability of deep networks to small, imperceptible perturbations was first demonstrated by \citet{szegedy2014intriguing} and explained via the linear-behavior hypothesis by \citet{goodfellow2015explaining}, who introduced the fast gradient sign method (FGSM). \citet{kurakin2017scale, kurakin2018adversarial} extended this to an iterative variant (I-FGSM/BIM), improving white-box attack strength at some cost to transferability, a trade-off later characterized more broadly by \citet{tramer2018ensemble}. Optimization-based attacks such as \citet{carlini2017towards} instead treat the perturbation bound as a soft constraint via Lagrangian relaxation. Momentum-based iterative attacks \citep{dong2018boosting}, as adapted by \citet{nie2025new}, were introduced primarily to improve black-box transferability by smoothing the gradient signal across iterations and escaping poor local optima. All of this work shares the same basic constraint structure: perturbations are bounded \emph{above} by a small $\epsilon$, and the attack succeeds when the prediction changes.

\textbf{Fooling images and inverted formulations.} \citet{nguyen2015deep} showed that evolutionary search can produce images that are unrecognizable to humans yet classified with high confidence by a trained network, establishing that high-confidence predictions do not imply the input resembles the training distribution in any way a human would recognize. In earlier work \citep{borji2022new}, we reformulated this as a targeted, gradient-based attack in which the ``target'' class is simply the model's own correct prediction on the original image, and introduced the constraint structure in Eq.~\ref{eq:nke}, framing it as the mirror image of the classical adversarial setting along the perturbation axis. \citet{nie2025new} developed this direction further with $L_2$-norm variants and momentum-based extensions, and are --- to our knowledge --- the first to evaluate this attack family at ImageNet scale and to systematically measure its transferability across architectures, finding it to be substantially worse than that of classical adversarial examples.

\textbf{Human comparison in adversarial robustness.} A separate line of work has compared human and model behavior under adversarial or corrupted inputs \citep{elsayed2018adversarial, geirhos2018generalisation, zhou2019humans}, generally finding that models and humans diverge in systematic ways under both time-limited and unlimited viewing conditions. This literature provides methodology for measuring human psychometric functions under controlled perturbation, but has not been applied to the large-perturbation, same-label setting studied here; the human-model gap central to our framing \citep{borji2022new} (illustrated only conceptually in Fig.~3 there) has not been empirically measured for this specific attack family in either source paper.

\textbf{Out-of-distribution detection and calibration.} A large body of work addresses the related but distinct problem of detecting when a model is given an input unlike its training distribution, or of ensuring that a model's confidence is well-calibrated to its accuracy \citep{hendrycks2017baseline, lee2018simple, liu2020energy, ovadia2019can}. This literature treats ``confident but wrong, or confident on nonsensical input'' as a first-class failure mode to be measured and mitigated, using tools such as Mahalanobis-distance-based detectors, energy-based OOD scores, and expected calibration error. Despite the clear conceptual overlap --- a large-$\epsilon_l$ adversarial example is, definitionally, an input far from the natural data manifold that nonetheless receives a confident, correct prediction --- neither our earlier work \citep{borji2022new} nor \citet{nie2025new} evaluate these examples against any OOD detection or calibration baseline.

\textbf{Adversarial defenses and robustness.} Adversarial training \citep{goodfellow2015explaining, madry2018towards} and its ensemble variants \citep{tramer2018ensemble} remain the most widely used defenses against classical adversarial examples, alongside certified approaches such as randomized smoothing \citep{cohen2019certified} and simple input-transformation defenses \citep{guo2018countering}. Whether robustness to small-$\epsilon$ perturbations transfers to robustness against large-$\epsilon_l$, same-label perturbations is, to our knowledge, an open question.

\section{Methods}
\label{sec:methods}

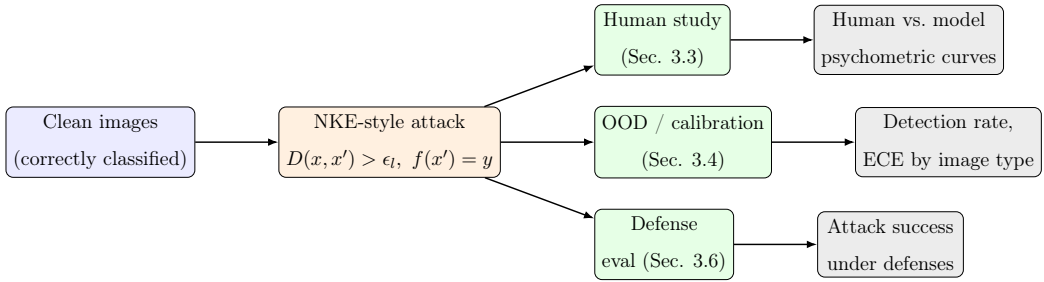
\begin{figure}[h]
\centering
\resizebox{\linewidth}{!}{%
\begin{tikzpicture}[
    node distance=1.1cm and 1.6cm,
    box/.style={draw, rounded corners, align=center, minimum height=1cm, minimum width=2.6cm, fill=gray!8, font=\small},
    arrow/.style={-{Latex[length=2mm]}, thick}
]
\node[box, fill=blue!8] (data) {Clean images\\(correctly classified)};
\node[box, right=of data, fill=orange!12] (attack) {NKE-style attack\\$D(x,x')>\epsilon_l,\ f(x')=y$};
\node[box, above right=0.6cm and 1.8cm of attack, fill=green!10] (human) {Human study\\(Sec. 3.3)};
\node[box, right=1.8cm of attack, fill=green!10] (ood) {OOD / calibration\\(Sec. 3.4)};
\node[box, below right=0.6cm and 1.8cm of attack, fill=green!10] (def) {Defense\\eval (Sec. 3.6)};
\node[box, right=1.6cm of human, fill=gray!15, minimum width=2.8cm] (out1) {Human vs.\ model\\psychometric curves};
\node[box, right=1.6cm of ood, fill=gray!15, minimum width=2.8cm] (out2) {Detection rate,\\ECE by image type};
\node[box, right=1.6cm of def, fill=gray!15, minimum width=2.8cm] (out3) {Attack success\\under defenses};
\draw[arrow] (data) -- (attack);
\draw[arrow] (attack) -- (human);
\draw[arrow] (attack) -- (ood);
\draw[arrow] (attack) -- (def);
\draw[arrow] (human) -- (out1);
\draw[arrow] (ood) -- (out2);
\draw[arrow] (def) -- (out3);
\end{tikzpicture}%
}
\caption{Overview of the experimental pipeline. A single set of NKE-style adversarial images, generated at multiple perturbation magnitudes $\epsilon_l$, feeds independent evaluation tracks whose outputs are analyzed jointly in Sec.~\ref{sec:results}.}
\label{fig:pipeline}
\end{figure}

Our study is organized around a single attack-generation pipeline that feeds three independent evaluation tracks plus a mechanistic analysis (Fig.~\ref{fig:pipeline}). A single set of NKE-style images, generated at multiple perturbation magnitudes $\epsilon_l$, feeds every track, so that all downstream analyses operate on identical stimuli.

\subsection{Attack generation}
\label{sec:attack-gen}
We generate adversarial examples satisfying Eq.~\ref{eq:nke} by minimizing the true-label loss (to keep the prediction correct) while an exterior projection holds the perturbation at $L_2$ distance $\ge \epsilon_l$ from the clean image:
\[
x'_{N+1} = \Pi_{\|\cdot-x\|_2 \ge \epsilon_l}\!\Big[\operatorname{clip}_{[0,1]}\big( x'_N - \alpha\, g_N \big)\Big],
\]
where $g_N$ is the (optionally momentum-accumulated, sign- or $L_2$-normalized) gradient of $J(x'_N,y)$. Here $J(\cdot,\cdot)$ is the model's classification loss (cross-entropy between the predicted and true class), evaluated at the current iterate $x'_N$ against the original true label $y$; minimizing it, rather than maximizing it as in a classical attack, is what keeps the model's prediction correct throughout the attack. $x'_0=x$ is the clean image, $N$ indexes the attack iteration, and $\alpha$ is the step size. $\operatorname{clip}_{[0,1]}(\cdot)$ clamps pixel values back into the valid image range after each gradient step. $\Pi_{\|\cdot-x\|_2\ge\epsilon_l}$ is the exterior projection: whenever a step would leave the iterate inside the $L_2$ ball of radius $\epsilon_l$ around $x$, it pushes the iterate back out to the ball's boundary, enforcing the reversed, lower-bound perturbation constraint of Eq.~\ref{eq:nke} (the mirror image of the usual upper-bound projection used in classical adversarial attacks). We implement our earlier plain iterative form \citep{borji2022new} and the $L_2$/momentum variants (NI-FGM, NMI-FGSM, NMI-FGM) of \citet{nie2025new} in a shared codebase, and use NMI-FGSM/NMI-FGM (60 steps) as the primary stimulus generator, as momentum variants are the most reliable \citep{nie2025new}. We restrict all tracks to images correctly classified before the attack and still classified with the original label after, per Eq.~\ref{eq:nke}. We use two architecturally distinct models per dataset: LeNet/MLP on MNIST and ResNet-18/VGG-11 on CIFAR-10 (clean test accuracy 99.3/98.5\% and 93.5/90.2\% respectively).

\subsection{Baseline replication}
\label{sec:baseline-replication}
Before the novel tracks, we validate our pipeline against the phenomena reported by \citet{nie2025new} at three scales: MNIST (LeNet, MLP), CIFAR-10 (ResNet-18, VGG-11), and \textbf{ImageNet} --- using the ImageNet-pretrained ResNet-152 and VGG-16 (two architecturally distinct families) on \emph{Imagenette}, a 10-class subset of real ImageNet images at $224\times224$. The full ImageNet training set is unavailable in our environment, but the NKE attack only requires correctly-classified natural images, which Imagenette provides; this lets us measure \citeauthor{nie2025new}'s ImageNet transfer collapse directly rather than cite it. We reproduce (i) near-perfect white-box success and (ii) the white-box$\gg$black-box asymmetry at all three scales (Table~\ref{tab:baseline-matrix}), and reuse \citeauthor{nie2025new}'s three ablation axes (perturbation size, iterations, momentum $\mu$) at coarser resolution.

\subsection{Human psychometric study}
\label{sec:human}
The central untested assumption, from our earlier work, is that human accuracy degrades faster than model accuracy as $\epsilon_l$ increases \citep[Fig.~3]{borji2022new}. As \citet{borji2022addressing} argues, $L_p$ distance is not a reliable proxy for perceptual similarity, so this must be validated rather than assumed from the $\epsilon_l$ bound. \textbf{Protocol (released as a harness).} Participants view one image at a time and select the category from the dataset's label set, with an explicit ``cannot tell'' option; each image is shown to multiple raters, no participant sees the same base image at two perturbation levels, and a matched-budget Gaussian-noise control at the same $L_2$ distance is included to separate adversarial structure from generic signal degradation \citep[addressing][]{borji2022addressing, elsayed2018adversarial, zhou2019humans}. We export the full stimulus set + a self-contained forced-choice task page. The harness enforces the design client-side: each participant is assigned a between-subjects sample so that no base image is seen at two perturbation levels, and the clean ($\epsilon_l{=}0$) images are interleaved as attention checks. Returned sessions are pooled and attention-filtered --- participants whose clean-trial accuracy falls below a dataset-calibrated threshold are discarded, as the clean-image human ceiling on $32\times32$ CIFAR-10 is well below MNIST's --- and we report human accuracy, ``cannot tell'' rate, and reaction time per $\epsilon_l$ for the NKE and Gaussian-control conditions, with bootstrap 95\% confidence intervals. \textbf{Proxies reported here.} In lieu of (not as a replacement for) crowd data, we report two labeled computational recognizability proxies: (a) a \emph{generic recognizer} --- the other architecture, trained independently and never exposed to the attack gradients; and (b) a \emph{shape recognizer} --- a classifier trained on Canny edge maps. These are proxies, not human data.

\subsection{Out-of-distribution and calibration analysis}
\label{sec:ood}
We evaluate three post-hoc OOD detectors --- maximum softmax probability (MSP) \citep{hendrycks2017baseline}, a Mahalanobis feature-distance detector \citep{lee2018simple}, and an energy score \citep{liu2020energy} --- reporting the fraction of NKE images flagged at a threshold calibrated to 5\% false-positive rate on \emph{held-out} clean data (disjoint from the split used to fit the detector; an in-sample threshold understates the true FPR for a fitted-covariance detector), alongside classical small-$\epsilon$ adversarials and genuine OOD (SVHN/FashionMNIST) as references. We also report the threshold-independent AUROC, and compute expected calibration error (ECE) \citep{ovadia2019can} for clean, classical-adversarial, and NKE images.

\subsection{Mechanistic analysis: shape vs.\ texture}
\label{sec:mechanism}
Standard CNNs are texture-biased whereas humans rely on shape \citep{geirhos2018imagenet, borji2022addressing}. We test whether the attack preferentially destroys texture while leaving edge/shape structure comparatively intact by computing, per image and $\epsilon_l$, an edge-preservation score (Canny-edge-map F1, clean vs.\ perturbed) and a texture-similarity score: we compute the GLCM (a matrix tallying how often pairs of pixel gray-levels co-occur at a fixed offset, from which standard texture statistics such as contrast, homogeneity, and energy are derived) separately for the clean and perturbed image, and report the cosine similarity between their GLCM feature vectors, so a score of 1 means texture statistics are unchanged and 0 means they are unrelated; plus, for comparison, a Gram-matrix cosine.

% and a texture-similarity score (gray-level co-occurrence matrix, GLCM, features; plus a Gram-matrix cosine).

\subsection{Robustness intervention evaluation}
\label{sec:defenses}
We re-run the white-box attack against (i) an adversarially trained model \citep{madry2018towards}, (ii) randomized smoothing \citep{cohen2019certified}, and (iii) input-transformation defenses (JPEG, Gaussian blur, random resize/pad) \citep{guo2018countering}, using straight-through gradients for non-differentiable transforms. We correlate each model's classical small-$\epsilon$ robust accuracy with its resistance to the large-$\epsilon_l$ NKE attack (operationalized as $1-\overline{\text{success}}$ over $\epsilon_l>0$).

\section{Results}
\label{sec:results}

\begin{figure}[t]
\centering
\includegraphics[width=\linewidth]{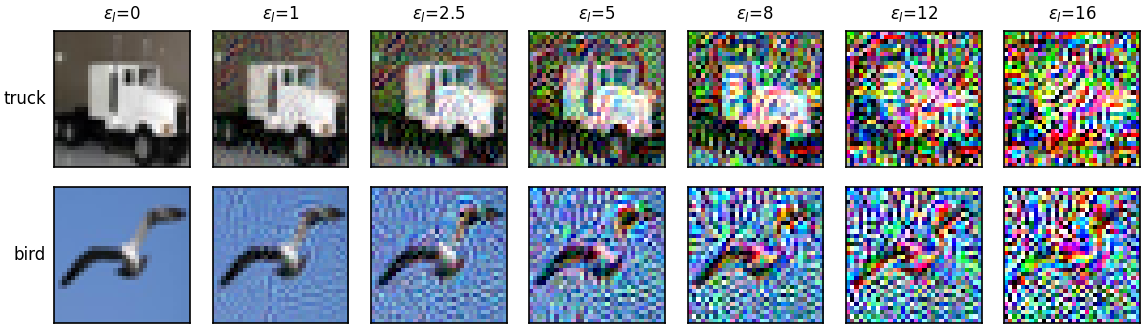}
\caption{NKE examples (CIFAR-10, ResNet-18): clean ($\epsilon_l{=}0$) $\to$ NKE across increasing $\epsilon_l$. The model keeps the original label at $\sim$1.0 confidence across each entire row, even where the image has dissolved into colour noise no human could label. Full grid in Fig.~\ref{fig:qual}.}
\label{fig:teaser}
\end{figure}

Across \emph{all} conditions below, the source model's accuracy on the perturbed images is 100\% at $\sim$1.0 confidence, out to perturbations as large as the image content itself ($L_2\approx 11$ on MNIST, $\approx 16$ on CIFAR-10). Figure~\ref{fig:teaser} shows what these images look like: recognizable objects dissolve into colour noise as $\epsilon_l$ grows, yet the model's label and confidence are unchanged along every row (the full grid, and an ImageNet analogue, are in Fig.~\ref{fig:qual}, App.~\ref{app:extra}). The NKE property is trivially satisfied; the interesting variation is in everything else.

\subsection{Baseline replication and the transferability--complexity trend}
\label{sec:res-baseline}
\begin{figure}[t]
\centering
\includegraphics[width=0.6\linewidth]{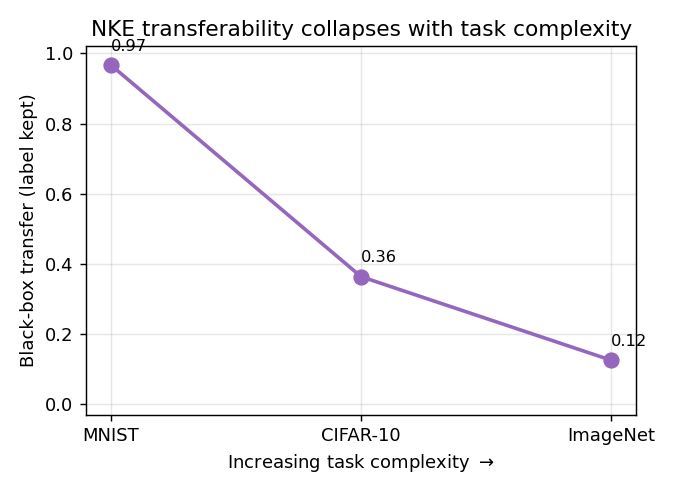}
\caption{Black-box transfer (label kept) vs.\ task complexity, measured on all three scales (ImageNet at $\epsilon_l{=}30$, $L_2\!\approx\!61$). Transferability collapses as complexity rises.}
\label{fig:trend}
\end{figure}

White-box success is $\approx$100\% and cross-model (black-box) success is far lower, replicating \citeauthor{nie2025new}'s asymmetry on CIFAR-10 (Table~\ref{tab:baseline-matrix}). Ablations are flat at 1.0 across perturbation size and $\mu\in[0,1]$, and across iterations $\ge 10$ (5 iterations $\to$ 0.875 on CIFAR-10). Across all three scales, measured on real data, a clean trend emerges (Fig.~\ref{fig:trend}): \textbf{NKE transferability falls as task complexity rises} --- MNIST $\approx$0.97 $\to$ CIFAR-10 $\approx$0.36 $\to$ ImageNet $\approx$0.10 (and $\to$0.00 at larger $\epsilon_l$), the last measured with ResNet-152/VGG-16 on Imagenette rather than cited. On simple data the perturbation preserves class-defining structure that generalizes across models; on complex data it becomes model-specific. Our simple-dataset regime~\cite{{borji2022new}} and \citeauthor{nie2025new}'s ImageNet regime are two ends of one continuum.

\begin{table}[t]
\centering
\small
\caption{Reduced-scale replication of the white-box/black-box asymmetry. Entries are the fraction of source-generated NKE images the target model still labels correctly (label-kept rate; $\epsilon_l$ at the second-largest grid level). $*$ = white-box. \citeauthor{nie2025new}'s ImageNet black-box transfer ($<$0.05) is shown as an external reference.}
\label{tab:baseline-matrix}
\begin{tabular}{l l c c}
\toprule
Dataset & Source model & Success vs.\ Model A & Success vs.\ Model B \\
\midrule
\multirow{2}{*}{MNIST}    & LeNet (A)   & 1.00* & 0.98 \\
                          & MLP (B)     & 0.95  & 1.00* \\
\midrule
\multirow{2}{*}{CIFAR-10} & ResNet-18 (A) & 1.00* & 0.50 \\
                          & VGG-11 (B)    & 0.23  & 1.00* \\
\midrule
\multirow{2}{*}{ImageNet}   & ResNet-152 (A) & 1.00* & 0.10 \\
                          & \multicolumn{3}{l}{\footnotesize (transfer $\to$0.01 at $\epsilon_l{=}60$, 0.00 at $\epsilon_l{=}100$; cf.\ \citeauthor{nie2025new} $<$0.05)} \\
\bottomrule
\end{tabular}
\end{table}

\subsection{The human--model gap is real on natural images (and, on the proxy, specific to adversarial structure)}
\label{sec:res-human}
\begin{figure}[t]
\centering
\includegraphics[width=0.9\linewidth]{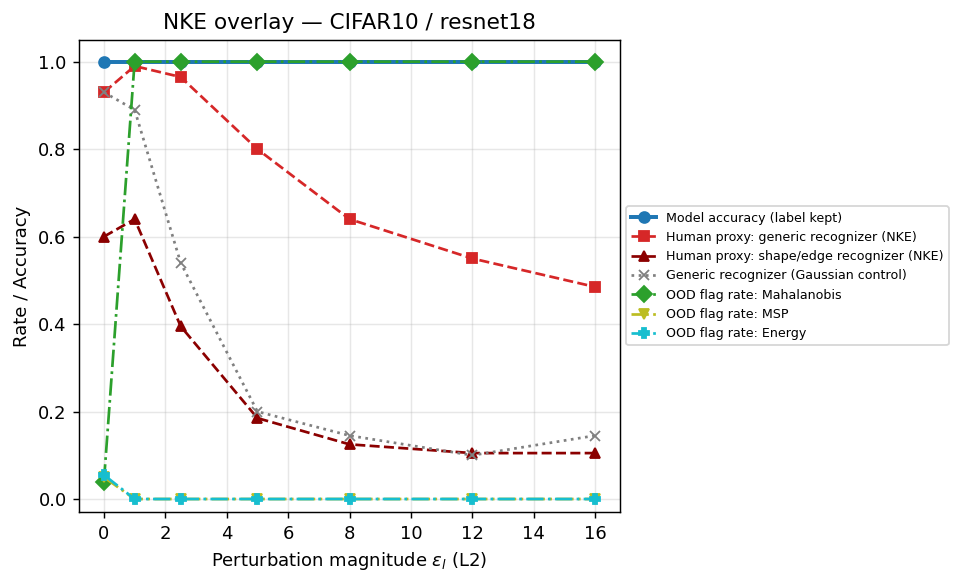}
\caption{Summary overlay (CIFAR-10, ResNet-18). Model accuracy (label kept) stays at 1.0; the generic and shape recognizer proxies (proxies for human accuracy) fall well below it; the matched Gaussian control falls fastest; MSP/energy OOD flag rates stay at $\approx$0 while Mahalanobis flags 100\%. The gap between the flat model curve and the descending proxy curves is the human--model gap that prior work only illustrated schematically.}
\label{fig:overlay}
\end{figure}

Table~\ref{tab:human} reports the proxies (summary overlay in Fig.~\ref{fig:overlay}). On CIFAR-10, the independent generic recognizer falls to \textbf{48.5\%} at $\epsilon_l=16$ while the source model stays at 100\% --- a $\sim$50-point gap, the first quantitative (proxy) evidence for the gap the original formulation assumed. Crucially, a \textbf{matched-$L_2$ Gaussian control degrades the generic recognizer faster} (to $\sim$0.15) than the structured NKE perturbation (to 0.49). The control is constructed per image and per $\epsilon_l$: we draw i.i.d.\ Gaussian noise, rescale it so its $L_2$ norm exactly equals that of the actual NKE perturbation generated for that image at that $\epsilon_l$, add it to the clean image, and clip to $[0,1]$ --- so every control image removes exactly as much signal, in the $L_2$ sense, as its NKE counterpart, differing only in whether the perturbation direction is random or adversarially structured. This directly addresses \citeauthor{borji2022addressing}'s confound concern that $L_p$ distance alone is not a valid proxy for perceptual similarity: holding magnitude fixed while varying only structure means any recognizability difference between the two conditions isolates the effect of adversarial \emph{structure} itself, not the amount of signal removed. That the random-direction control is in fact \emph{more} destructive is, on reflection, unsurprising: the gradient-based attack must also keep $f(x')=y$ (Eq.~\ref{eq:nke}), so part of its perturbation budget is spent moving along directions the classifier's decision boundary tolerates, which are not the directions that maximally destroy human-recognizable structure, whereas isotropic Gaussian noise carries no such constraint and spends its entire budget indiscriminately. The shape recognizer collapses toward chance. On MNIST the generic proxy stays high (0.99$\to$0.94) --- NKE digits remain broadly recognizable (mirroring the high MNIST transfer above) --- so the gap is a natural-image phenomenon.

\textbf{Empirical human curve (pilot, $N{=}5$).} Using the released, hardened harness we collected real forced-choice responses under the protocol of Sec.~\ref{sec:human}. Table~\ref{tab:human-real} reports a \emph{pilot} of $N{=}5$ attention-passing participants (all five collected sessions passed the dataset-calibrated attention check; 260 judgements retained). \textbf{This is a small-sample pilot, not a definitive human study}: per-cell counts are modest (12--23 per $\epsilon_l$) and the bootstrap 95\% CIs correspondingly wide. Even so, the predicted pattern is stark: human recognition of NKE images falls from near-ceiling (0.88 at $\epsilon_l{=}2.5$) to \textbf{$\sim$0.08 by $\epsilon_l{=}12$--$16$}, with ``cannot tell'' selected on \textbf{78--83\%} of trials at the largest perturbations, while model accuracy stays at 100\% --- a human--model gap widening to $\sim$0.92, direct human evidence for the gap our original formulation only illustrated schematically \citep[Fig.~3]{borji2022new}. The human ceiling is already only 0.85 on the \emph{clean} $32\times32$ images, confirming that a dataset-calibrated attention threshold is necessary (our CIFAR-10 gate of 0.55 is set from this clean-trial ceiling, not tuned to retain participants). The matched-$L_2$ Gaussian contrast is more equivocal than on the proxy: raters find NKE \emph{more} recognizable than equal-energy noise at small-to-mid $\epsilon_l$ (\eg 0.63 vs.\ 0.50 at $\epsilon_l{=}5$), the same direction as the proxy's ``not signal loss'' effect, but the conditions are \emph{not} statistically separable at this sample size (overlapping CIs). Establishing the signal-vs-structure dissociation on real humans therefore requires the properly-powered sample the harness is built to collect; for that claim the proxy (Table~\ref{tab:human}) remains our primary evidence, while the pilot corroborates the gap itself.
% Populated by nke.exp_human_analysis from results/human_real_cifar10_resnet18.json (N=5 pilot; all 5 collected sessions used, no further participants available).
\begin{table}[t]
\centering
\small
\caption{\textbf{Pilot ($N{=}5$ attention-passing participants) --- a small-sample study.} Empirical human forced-choice accuracy vs.\ $\epsilon_l$ (CIFAR-10, ResNet-18 stimuli); all five collected sessions passed the dataset-calibrated attention check, 260 judgements retained. Per-cell counts (last row) are modest, so the bootstrap 95\% CIs are wide. Model accuracy is 1.00 at every level. ``Gauss'' = matched-$L_2$ Gaussian control; the $\epsilon_l{=}0$ column is the clean baseline (shared by both conditions).}
\label{tab:human-real}
\begin{tabular}{l ccccccc}
\toprule
$\epsilon_l$ & 0 & 1 & 2.5 & 5 & 8 & 12 & 16 \\
\midrule
Human accuracy (NKE)          & 0.85 & 0.88 & 0.88 & 0.63 & 0.37 & 0.09 & 0.08 \\
\quad bootstrap 95\% CI       & \footnotesize[.72,.95] & \footnotesize[.69,1.0] & \footnotesize[.71,1.0] & \footnotesize[.42,.84] & \footnotesize[.16,.58] & \footnotesize[.00,.22] & \footnotesize[.00,.25] \\
Human accuracy (Gauss ctrl)   & 0.85 & 0.73 & 0.81 & 0.50 & 0.44 & 0.08 & 0.05 \\
``Cannot tell'' rate (NKE)    & 0.07 & 0.06 & 0.00 & 0.11 & 0.37 & 0.78 & 0.83 \\
Human--model gap (NKE)        & 0.15 & 0.12 & 0.12 & 0.37 & 0.63 & \textbf{0.91} & \textbf{0.92} \\
\# judgements (NKE)           & 40   & 16   & 17   & 19   & 19   & 23   & 12 \\
\bottomrule
\end{tabular}
\end{table}

\begin{table}[t]
\centering
\small
\caption{Recognizability proxies vs.\ $\epsilon_l$ (CIFAR-10, ResNet-18 source; generic recognizer = VGG-11). Model accuracy is 1.00 at every level. ``Gauss'' = matched-$L_2$ Gaussian control.}
\label{tab:human}
\begin{tabular}{l ccccc}
\toprule
$\epsilon_l$ & 0 & 2.5 & 5 & 8 & 16 \\
\midrule
Generic recognizer (NKE)       & 0.93 & 0.965 & 0.80 & 0.64 & \textbf{0.485} \\
Generic recognizer (Gauss ctrl)& 0.93 & 0.54  & 0.20 & 0.15 & 0.145 \\
Shape recognizer (NKE)         & 0.60 & 0.40  & 0.19 & 0.13 & 0.105 \\
\bottomrule
\end{tabular}
\end{table}

\subsection{NKE is an OOD phenomenon invisible to confidence-based monitoring}
\label{sec:res-ood}
Table~\ref{tab:ood} shows a sharp detector dissociation. On CIFAR-10, MSP and energy detectors flag \textbf{0\%} of NKE images and ECE is $\approx$0 --- the model is confident \emph{and} correct, so calibration looks perfect on inputs no human could label. The \textbf{Mahalanobis feature-distance detector flags 100\%} of them (from $\epsilon_l=1$), at a threshold calibrated to a true 5\% FPR on held-out clean data; its AUROC vs.\ clean is $\approx$1.00 at every $\epsilon_l\ge1$, so the detection reflects score separation, not threshold choice. NKE images are far off-manifold in representation space yet indistinguishable at the output layer --- arguing for feature-distance OOD detection over softmax/energy/temperature methods when such failures matter, a reframing neither source paper performed, though this advantage holds only against an attacker \emph{oblivious} to the detector (Sec.~\ref{sec:res-adaptive}).

\begin{table}[t]
\centering
\small
\caption{OOD detection rate at a held-out-calibrated 5\% FPR, and ECE, for NKE images (CIFAR-10, ResNet-18). References: genuine OOD (SVHN) $\to$ Maha 0.70 / MSP 0.53; classical PGD adversarials $\to$ Maha 0.98 / MSP 0.00. NKE is thus flagged by Mahalanobis \emph{more} reliably (1.00) than genuine OOD.}
\label{tab:ood}
\begin{tabular}{l ccccc}
\toprule
$\epsilon_l$ & 1 & 2.5 & 5 & 8 & 16 \\
\midrule
Model accuracy      & 1.00 & 1.00 & 1.00 & 1.00 & 1.00 \\
ECE                 & 0.00 & 0.00 & 0.00 & 0.00 & 0.00 \\
MSP detection       & 0.00 & 0.00 & 0.00 & 0.00 & 0.00 \\
Energy detection    & 0.00 & 0.00 & 0.00 & 0.00 & 0.00 \\
\textbf{Mahalanobis detection} & \textbf{1.00} & \textbf{1.00} & \textbf{1.00} & \textbf{1.00} & \textbf{1.00} \\
\bottomrule
\end{tabular}
\end{table}

\subsection{Feature-distance detection is not adaptively robust}
\label{sec:res-adaptive}
\begin{figure}[t]
\centering
\includegraphics[width=0.72\linewidth]{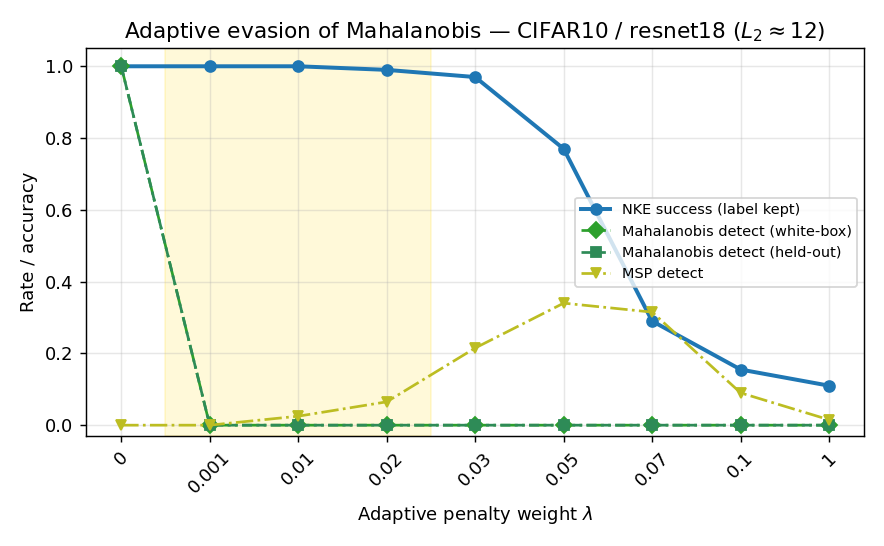}
\caption{Adaptive evasion of the Mahalanobis detector (CIFAR-10, ResNet-18, $L_2{\approx}12$). A tiny penalty $\lambda$ drives detection (white-box and held-out, green) to 0\% while NKE success (blue) stays $\ge$0.99 across the shaded band; only much larger $\lambda$ costs success. MSP detection (yellow) never recovers the gap.}
\label{fig:adaptive}
\end{figure}

The Mahalanobis result above assumes an attacker oblivious to the detector. But feature distance is differentiable, so an adaptive attacker can fold it into the objective: minimise $\mathrm{CE}(f(x'),y) + \lambda\, d_{\mathrm{Maha}}(x')$ --- keeping the label \emph{and} pulling the penultimate feature back toward the clean class manifold --- while still projecting the perturbation to $\|x'-x\|_2 \ge \epsilon_l$. 
Table~\ref{tab:adaptive} (plotted in Fig.~\ref{fig:adaptive}) sweeps $\lambda$ at a large $\epsilon_l$ ($L_2\approx12$) where oblivious NKE is flagged 100\%. A tiny penalty drives \textbf{Mahalanobis detection from 100\% to 0\%} while success stays $\ge$\textbf{0.99} out to $\lambda=0.02$ and the perturbation stays large ($L_2\approx12.06$, essentially unchanged); MSP stays $\le 0.07$. Only for larger $\lambda$, where the feature-pullback term overwhelms the large-$L_2$ projection, does success degrade (0.77 at $\lambda=0.05$, $\approx$0.11 by $\lambda=1$). Crucially, \textbf{the evasion is not white-box overfitting}: a \emph{held-out} detector fit on disjoint data is evaded identically, so the attacker need only know that a feature-distance detector is in use, not its parameters. There is thus a wide $\lambda$ band ($10^{-3}$ to $2{\times}10^{-2}$) that is simultaneously fully successful, large-perturbation, and fully evasive against both detectors: \textbf{feature-distance detection defeats a static NKE adversary but is defeated at no cost by an adaptive one}. The honest reframing is therefore not ``deploy Mahalanobis'' but that NKE is an off-manifold phenomenon which \emph{only} off-manifold detectors can see at all --- and even those are not robust to an adversary who knows they are there.

\begin{table}[t]
\centering
\small
\caption{Adaptive attack against the Mahalanobis detector (CIFAR-10, ResNet-18, at an $\epsilon_l$ with $L_2\approx12$, where oblivious NKE is 100\% detected). The attacker minimises $\mathrm{CE}+\lambda\,d_{\mathrm{Maha}}$; detection is at a held-out-calibrated 5\% FPR. Both a \emph{white-box} detector (the exact one attacked) and a \emph{held-out} detector (fit on disjoint data) are evaded to 0\% across a wide $\lambda$ band, at no cost to success or perturbation magnitude.}
\label{tab:adaptive}
\begin{tabular}{l cccccccc}
\toprule
$\lambda$ & 0 & 0.001 & 0.01 & 0.02 & 0.03 & 0.05 & 0.1 & 1.0 \\
\midrule
NKE success                      & 1.00 & 1.00 & 1.00 & 0.99 & 0.97 & 0.77 & 0.16 & 0.11 \\
Perturbation $L_2$               & 12.02 & 12.05 & 12.05 & 12.06 & 12.07 & 12.06 & 12.07 & 12.06 \\
\textbf{Maha detection (white-box)} & \textbf{1.00} & \textbf{0.00} & \textbf{0.00} & \textbf{0.00} & 0.00 & 0.00 & 0.00 & 0.00 \\
\textbf{Maha detection (held-out)}  & \textbf{1.00} & \textbf{0.00} & \textbf{0.00} & \textbf{0.00} & 0.00 & 0.00 & 0.00 & 0.00 \\
MSP detection                    & 0.00 & 0.00 & 0.03 & 0.07 & 0.21 & 0.34 & 0.09 & 0.01 \\
\bottomrule
\end{tabular}
\end{table}

Two further checks (full detail in App.~\ref{app:extra}, Table~\ref{tab:detectors}) confirm this is a property of feature-distance detection in general, not of one metric or model. First, the evasion generalises across detector \emph{design}: under honest held-out threshold calibration, a two-layer ensemble objective drives \textbf{every} parametric detector we tested --- tied- and per-class-covariance, at both layers --- to 0\% detection at 100\% success, while a non-parametric cosine-$k$NN detector \citep{sun2022knn} never flags standard NKE at all (a per-class detector only \emph{appears} robust under in-sample calibration, an artifact of its 77\% true FPR --- a calibration caution in its own right). Second, the no-cost small-$\lambda$ evasion reproduces against the adversarially trained ResNet-18 (45\% PGD robust accuracy), consistent with the orthogonality of NKE and classical robustness.

\subsection{Mechanism: texture is destroyed far faster than shape}
\label{sec:res-mech}
\begin{figure}[t]
\centering
\includegraphics[width=0.6\linewidth]{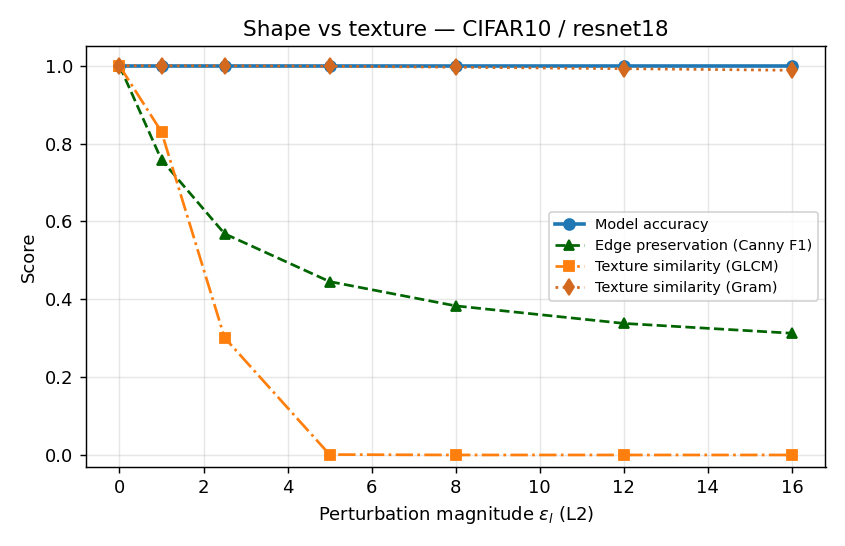}
\caption{Shape vs.\ texture dissociation (CIFAR-10, ResNet-18). Model accuracy (blue) stays at 1.0 while GLCM texture similarity (orange) collapses to 0 by $\epsilon_l{=}5$; edge preservation (green) survives far longer. The Gram-matrix cosine (red) is dominated by global colour energy and stays $\approx$1 throughout --- uninformative, unlike GLCM.}
\label{fig:mechanism}
\end{figure}

Table~\ref{tab:mech} (plotted in Fig.~\ref{fig:mechanism}) shows a channel dissociation: GLCM \textbf{texture similarity collapses to $\approx$0 by $\epsilon_l=5$}, while \textbf{edge/shape structure survives far longer} (F1 plateaus at $\sim$0.31--0.45), with model accuracy pinned at 1.00. The surviving edge skeleton is insufficient for the shape/generic recognizers to keep up, yet the source model is invariant to both channels --- consistent with a texture/shape account of the gap \citep{geirhos2018imagenet}. (The Gram-matrix cosine stayed $\sim$1.0 throughout; it is dominated by global colour energy and is uninformative --- GLCM is the meaningful texture metric.)

\begin{table}[t]
\centering
\small
\caption{Shape vs.\ texture (CIFAR-10, ResNet-18): mean clean-vs-perturbed edge preservation and texture similarity by $\epsilon_l$.}
\label{tab:mech}
\begin{tabular}{l ccccc}
\toprule
$\epsilon_l$ & 1 & 2.5 & 5 & 8 & 16 \\
\midrule
Edge preservation (Canny F1) & 0.76 & 0.57 & 0.45 & 0.38 & 0.31 \\
Texture similarity (GLCM)    & 0.83 & 0.30 & \textbf{0.00} & 0.00 & 0.00 \\
\bottomrule
\end{tabular}
\end{table}

\subsection{Architecture generality: does shape bias close the gap? (a Vision Transformer)}
\label{sec:res-vit}
Sec.~\ref{sec:res-mech} shows the attack spares shape while destroying texture, and CNNs are texture-biased whereas humans --- and \emph{vision transformers} --- are comparatively shape-biased \citep{geirhos2018imagenet, naseer2021intriguing, tuli2021convolutional}, yielding a falsifiable prediction: a ViT should read the surviving edge skeleton \emph{better} than a second CNN does. We add a small from-scratch ViT \citep{dosovitskiy2021image} (patch~4, dim~192, depth~6; 1.8M params, 83.2\% clean acc.) as a third CIFAR-10 architecture and re-run every track (full per-track tables and figure in App.~\ref{app:vit}). Three findings result. \emph{(i) The mechanism is architecture-general:} ViT-sourced NKE reproduces the Sec.~\ref{sec:res-mech} texture-before-shape dissociation exactly (GLCM $\to\approx$0 by $\epsilon_l{=}5$, edge F1 plateauing at $\sim$0.32--0.44), so it is not a CNN artifact. \emph{(ii) But shape bias does not close the gap (prediction falsified):} comparing retention (accuracy on NKE $\div$ on that source's clean images, removing the clean-accuracy confound), the shape-biased ViT recognizer retains \textbf{0.08--0.11 \emph{less}} than the second CNN on CNN-sourced NKE at every $\epsilon_l\ge2.5$ --- the surviving edge skeleton does not let even a shape-biased reader recover the label, so the human--model gap is not a texture-bias artifact. \emph{(iii) Attention models produce the most model-specific NKE:} mean cross-model retention at $\epsilon_l{=}16$ is 0.41 (ResNet-18 source), 0.21 (VGG-11), and \textbf{0.15} (ViT); the generic recognizer of ViT-sourced NKE falls to 0.12--0.14, barely above its matched-Gaussian control (0.10), the human--model gap is largest here ($\sim$86 points), and on the ViT penultimate the Mahalanobis detector catches only 43--50\% of \emph{oblivious} NKE (vs.\ 100\% on ResNet-18) while the adaptive attack still evades at $\lambda{=}0.001$ --- so the feature-distance detection advantage of Sec.~\ref{sec:res-ood} is itself architecture-dependent and favourable to CNNs.

\subsection{No classical defense mitigates NKE; robustness is orthogonal}
\label{sec:res-def}
\begin{figure}[t]
\centering
\includegraphics[width=\linewidth]{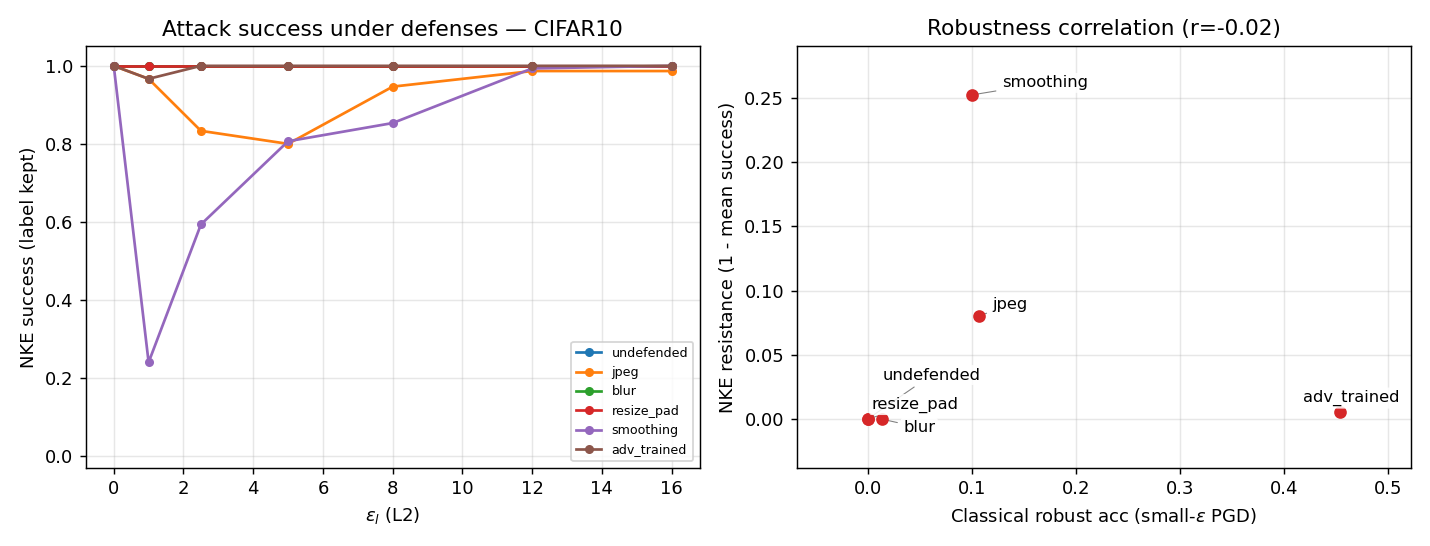}
\caption{\textbf{Left:} NKE success stays $\approx$1.0 across $\epsilon_l$ for every defense (CIFAR-10, ResNet-18 base); only randomized smoothing dips briefly at small $\epsilon_l$ before recovering. \textbf{Right:} classical robust accuracy (PGD) vs.\ NKE resistance ($1-\overline{\text{success}}$) across defenses, $r\approx-0.02$ --- robustness to small-$\epsilon$ attacks does not predict resistance to large-$\epsilon_l$ NKE.}
\label{fig:defenses}
\end{figure}

Table~\ref{tab:def} (plotted in Fig.~\ref{fig:defenses}) shows that every defended model remains $\approx$100\% NKE-vulnerable, including an adversarially trained model with \textbf{45\% PGD robust accuracy}. Across models, the correlation between classical robust accuracy and continuous NKE resistance is \textbf{$r\approx-0.02$} on CIFAR-10 (and undefined-with-zero-variance on MNIST, where NKE resistance is uniformly 0 across a 0$\to$86\% robust-accuracy range --- the strongest possible statement of orthogonality). Resistance to small-$\epsilon$ attacks does not transfer to large-$\epsilon_l$, same-label attacks; they are distinct vulnerabilities. (Post-hoc blur/smoothing crater clean accuracy on this non-robust base model, confounding their rows; the adv-trained-vs-undefended contrast is the load-bearing comparison.)

\begin{table}[t]
\centering
\small
\caption{Defenses vs.\ NKE (CIFAR-10, ResNet-18 base). NKE success is measured white-box against each defended model at the largest $\epsilon_l$.}
\label{tab:def}
\begin{tabular}{l ccc}
\toprule
Model & Clean acc & Classical robust acc (PGD) & NKE success @ max $\epsilon_l$ \\
\midrule
Undefended            & 1.00 & 0.00 & 1.00 \\
JPEG ($q{=}40$)       & 0.78 & 0.11 & 0.99 \\
Random resize+pad     & 0.85 & 0.00 & 1.00 \\
Adversarial training  & 0.74 & \textbf{0.45} & \textbf{1.00} \\
\bottomrule
\end{tabular}
\end{table}

\subsection{Extensions based on \citet{nie2025new}}
\label{sec:res-ext}
\citeauthor{nie2025new} compared their four attack variants and studied transferability, both only through the lens of \emph{attack success rate}; we revisit both through our new metrics (full detail, Table~\ref{tab:variants} and Fig.~\ref{fig:transfer}, in App.~\ref{app:extra}). \emph{First}, projecting all four variants onto the same $L_2$ sphere ($L_2\approx8$), so they differ only in step rule and momentum, their reported success-rate gaps largely vanish (all $\ge$99.5\% success, all flagged 100\% by Mahalanobis and 0--3\% by MSP, near-identical edge preservation); only the human proxy varies modestly. \textbf{The OOD dissociation and confident-and-correct property are thus properties of the phenomenon, not the variant} --- the variant mainly trades off perturbation magnitude, which in turn drives the human gap. \emph{Second}, source-model manifold-proximity (negative Mahalanobis distance) predicts \emph{which} examples transfer only at small $\epsilon_l$ (AUROC 0.73 at $\epsilon_l{=}2.5$), decaying to chance and reversing by large $\epsilon_l$ (0.39 at $\epsilon_l{=}16$), while edge preservation never predicts transfer ($|r|\le0.12$). In the large-$\epsilon_l$ regime that defines NKE, transfer is disconnected from both manifold-distance and shape --- a quantitative deepening of \citeauthor{nie2025new}'s ``highly model-specific'' conclusion: the surviving signal is idiosyncratic to the source model, not a shared, human-aligned remnant.

\subsection{ImageNet scale: all findings hold with the models \citet{nie2025new} used}
\label{sec:res-imagenet}
We run the full pipeline with ImageNet-pretrained ResNet-152 (source) and VGG-16 (target) on real Imagenette images (Fig.~\ref{fig:imagenet}, App.~\ref{app:extra}). \emph{Most} findings reproduce at scale: white-box success and confidence stay at 1.00; black-box transfer collapses $0.81\to0.13\to0.01\to0.00$; Mahalanobis flags $\sim$100\% while MSP and energy stay at 0\%; and texture similarity collapses to 0 while edge preservation degrades only to $\sim$0.21. The transfer curve doubles as the generic-recognizer human proxy, so the human--model gap, the OOD dissociation, and the transfer collapse are all visible on one axis, all with the exact model class \citeauthor{nie2025new} studied.

\textbf{The cross-model proxy saturates at ImageNet --- but CLIP resolves it.} The second-CNN proxy above conflates ``a human would recognize this'' with ``this transfers to another CNN''; at ImageNet the NKE transfer collapse drives the proxy to its floor, so NKE (0.125 at $\epsilon_l=30$) and the matched-Gaussian control (0.155) become indistinguishable --- not because the gap is signal loss, but because the proxy has no headroom. We therefore add a \emph{stronger, non-CNN} recognizability proxy: zero-shot classification by CLIP (ViT-B/32, LAION-2B), which was never trained on these tasks and is not the attack's transfer target (Sec.~\ref{sec:res-clip}, Fig.~\ref{fig:clip}). CLIP has ample headroom and reveals a clear human--model gap at \emph{both} scales, confirming the effect is not merely a CIFAR artifact.

\subsection{CLIP as a stronger recognizability proxy}
\label{sec:res-clip}
We report CLIP zero-shot accuracy on NKE and matched-Gaussian-control images at both scales (Fig.~\ref{fig:clip}, App.~\ref{app:extra}; chance $=0.10$). Two things stand out. First, \textbf{a large human--model gap is confirmed on an independent, broadly-trained proxy at both scales}: while the source model stays at 100\%, CLIP falls to 0.24 (CIFAR-10) and 0.33 (ImageNet) at the largest $\epsilon_l$. Second, the NKE-vs-Gaussian relationship is \textbf{scale-dependent}: on CIFAR-10, NKE remains \emph{more} CLIP-recognizable than matched noise (\eg $\epsilon_l=8$: 0.66 vs.\ 0.17), replicating the ``not signal loss'' effect on a non-CNN proxy; on ImageNet the ordering \emph{reverses} --- NKE is \emph{less} recognizable than matched noise ($\epsilon_l=100$: 0.33 vs.\ 0.61), so the perturbation that keeps ResNet-152 confident actively destroys broad semantic content \emph{faster} than equal-energy noise: the examples are adversarial in the human direction too. Either way the gap is not explained by generic signal degradation --- on CIFAR NKE preserves more human-readable signal than noise, on ImageNet it destroys more --- but in both cases the model is unmoved.

\section{Discussion and Conclusion}
We answered the three questions the NKE formulation left open. \emph{(i)} The human--model gap is real on natural images (a $\sim$50-point proxy gap on CIFAR-10) and, on CIFAR-10, is \emph{specific to adversarial structure} rather than signal loss; a CLIP zero-shot proxy confirms the gap at ImageNet scale as well (there NKE is even \emph{more} destructive to broad semantics than equal-energy noise). \emph{(ii)} NKE is best understood as an OOD phenomenon invisible to confidence/energy/calibration monitoring and visible only to feature-distance detection --- and even that visibility is not adaptively robust: an attacker who adds the Mahalanobis distance to the objective evades detection (100\%$\to$0\%) at no cost to success or perturbation size, including on a held-out detector, an adversarially trained model, and (via a two-layer ensemble objective under honest held-out calibration) every feature-distance detector we tested. \emph{(iii)} It is orthogonal to classical adversarial robustness, so existing defenses do not address it. The mechanistic dissociation --- texture destroyed far faster than shape --- offers a candidate explanation linking all three findings; a Vision Transformer probe shows it is architecture-general, but --- against the natural shape-bias prediction --- a more shape-biased model is \emph{not} a better reader of the surviving structure, and attention models in fact generate the most model-specific NKE of all (Sec.~\ref{sec:res-vit}). Extending \citet{nie2025new}, we further show their variant-success differences vanish at matched perturbation magnitude, and their transferability collapse corresponds to a regime in which surviving signal is idiosyncratic to the source model rather than a shared, on-manifold remnant. The main limitation is that our primary human curve is a computational proxy; a small real-human pilot ($N{=}5$) corroborates the gap directly (human recognition $\to$$\sim$0.08 at large $\epsilon_l$ while the model stays at 1.00) but is underpowered for the signal-vs-structure contrast, for which the harness is built to collect a larger sample. The ImageNet-scale results use Imagenette (10 real ImageNet classes) rather than the full 1000-class validation set, which was unavailable in our environment. 

More broadly, NKE occupies a regime that existing evaluation practice does not cover: it is neither a small, imperceptible perturbation nor a naturally out-of-distribution input, yet it produces exactly the kind of confident-but-unreliable prediction that robustness and OOD evaluations exist to catch. Any system that leans on softmax confidence or energy scores as a first line of defense against distributional failure should be audited against this regime specifically, since Sec.~\ref{sec:res-ood} shows those tools are structurally blind to it, and Sec.~\ref{sec:res-adaptive} shows that even the one detector family that does see it stops seeing it as soon as the attacker knows it is there. This suggests at least three directions beyond the present study: scaling the released human-study harness from the $N{=}5$ pilot here to a properly powered, crowd-sourced sample; extending the analysis beyond vision to modalities such as audio and text, where a large, label-preserving perturbation is less obvious to construct but plausibly still exists; and pursuing detectors that are adaptively robust by construction, rather than only in a static, non-adversarial evaluation, since our results indicate that off-manifold detection alone is not a sufficient defense once it is a known target. We release the full pipeline, stimuli, and harness so that this regime can be audited as routinely as classical adversarial robustness already is. Code, figures, and the harness are available at \url{https://github.com/alikayyam/NKE_attack.git}.

{\small
\bibliographystyle{elsarticle-num-names}
\bibliography{references}
}

\appendix

\section{Human-study interface}
\label{app:interface}
\begin{figure}[h]
\centering
\includegraphics[width=\linewidth]{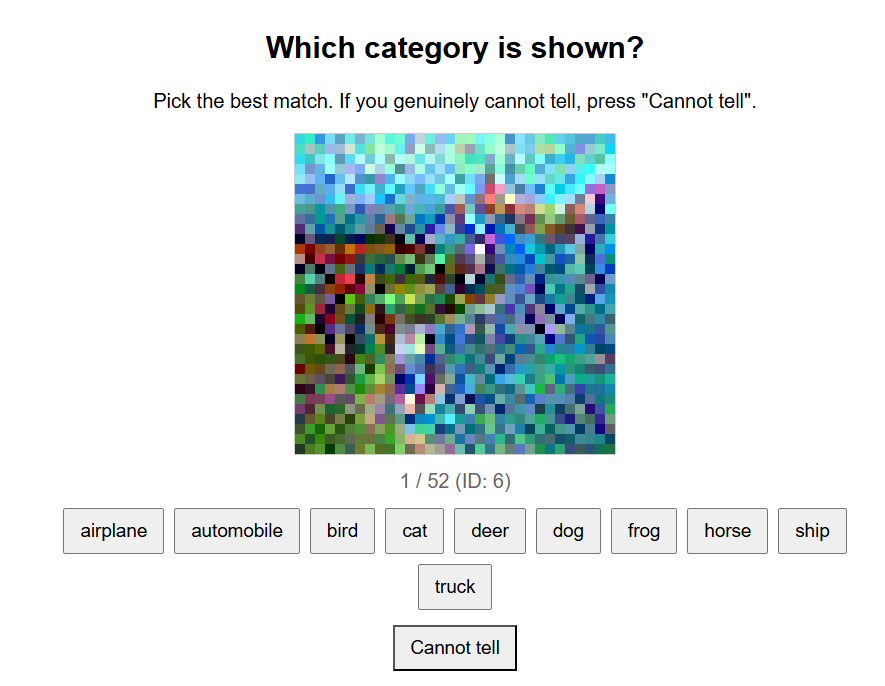}
\caption{The participant-facing forced-choice interface (CIFAR-10 stimuli shown). Each trial presents one enlarged, nearest-neighbor-scaled stimulus with a progress counter, ten class-label buttons, and an explicit \emph{Cannot tell} option.}
\label{fig:interface}
\end{figure}

This appendix describes the participant-facing interface of the released forced-choice harness (Sec.~\ref{sec:human}); Fig.~\ref{fig:interface} shows a representative screen.

\textbf{Layout.} Each trial shows a single stimulus image, enlarged to a fixed $256\times256$ on-screen size using nearest-neighbor (pixelated) scaling regardless of the underlying image's native resolution ($28\times28$ for MNIST, $32\times32$ for CIFAR-10), so no interpolation smoothing is introduced and every participant views the same effective magnification. Below the image, a progress indicator shows the current trial number out of the participant's total session length. Below that, one button per class label is shown (ten buttons for both MNIST and CIFAR-10, matching the number of dataset classes), followed by a single, visually distinct \emph{Cannot tell} button.

\textbf{Interaction.} The task begins with a participant-ID entry (auto-generated if left blank); no other setup or practice phase is required. On each trial, the participant clicks exactly one button --- a class label or Cannot tell --- which immediately advances to the next trial; there is no time limit and no way to revisit a previous trial. Response time is measured as wall-clock time between stimulus onset and the click. At the end of the session, a thank-you screen appears with a one-click download of the participant's own response log (or automatic submission, if a collection endpoint is configured).

\textbf{Trial composition and blinding.} Every clean ($\epsilon_l=0$) image is shown to every participant, serving as an interleaved attention check; these are visually indistinguishable from NKE and Gaussian-control trials, so participants cannot tell which condition a given trial belongs to. For the non-clean trials, each base image is shown to a given participant at exactly one perturbation level and under exactly one condition (NKE or Gaussian control), so no participant ever sees the same underlying image twice --- this is what makes the design between-subjects rather than within-subjects for the perturbed conditions, ruling out image-specific memory effects across conditions.

\textbf{Design rationale.} The explicit Cannot tell option is included because a standard $k$-way forced choice without an opt-out would conflate genuine recognition with a lucky guess at floor ($1/k$) once perturbation destroys most information; separating the two is necessary to interpret the low end of the human psychometric curve (Sec.~\ref{sec:res-human}) as reflecting inability to recognize the image rather than an unlucky guess. Fixed-size, nearest-neighbor display was chosen over native-resolution or interpolated display so that comparisons across $\epsilon_l$ and across datasets are not confounded by incidental differences in on-screen size or by smoothing artifacts that would themselves alter recognizability.

\section{Extended results and figures}
\label{app:extra}

\begin{figure}[h]
\centering
\includegraphics[width=\linewidth]{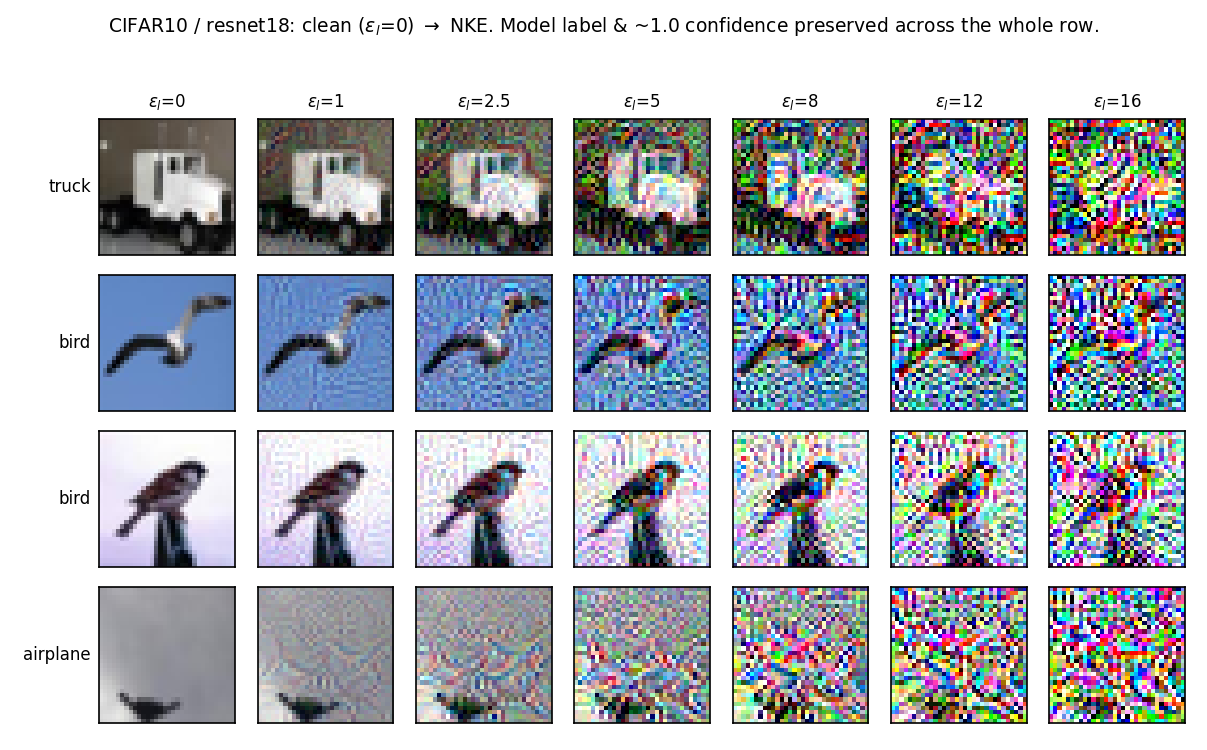}
\caption{CIFAR-10 examples: clean ($\epsilon_l=0$) $\to$ NKE across increasing $\epsilon_l$. ResNet-18 keeps the original label at $\sim$1.0 confidence across each entire row, even where the image is visually pure noise. (ImageNet analogue in the repository.)}
\label{fig:qual}
\end{figure}

\begin{figure}[h]
\centering
\includegraphics[width=0.62\linewidth]{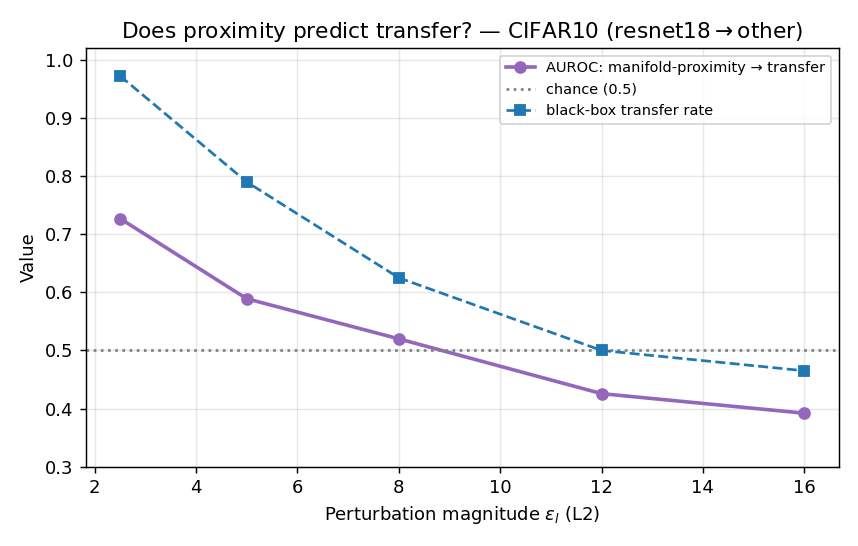}
\caption{Manifold-proximity predicts transfer only at small $\epsilon_l$; its predictive power (AUROC, purple) decays through chance and reverses as $\epsilon_l$ grows, tracking the black-box transfer rate (blue). In the NKE regime, transfer is not explained by proximity or shape.}
\label{fig:transfer}
\end{figure}

This appendix collects material deferred from the main text: the qualitative figure (Fig.~\ref{fig:qual}), the full feature-distance detector battery (Sec.~\ref{sec:res-adaptive}, Table~\ref{tab:detectors}), and the extended analysis of \citeauthor{nie2025new}'s variants and transferability (Sec.~\ref{sec:res-ext}, Table~\ref{tab:variants}, Fig.~\ref{fig:transfer}).

\subsection{Feature-distance detector battery (Sec.~\ref{sec:res-adaptive})}
Is the adaptive evasion specific to the tied-covariance penultimate detector, or does it generalise across detector \emph{design}? Holding $\epsilon_l$ fixed, we test a battery of feature-distance detectors, each calibrated to a true 5\% FPR on held-out clean data (Table~\ref{tab:detectors}). \emph{This last point is load-bearing:} with an in-sample threshold (calibrated on the detector's own fit set) a per-class (untied) covariance detector \emph{appeared} adaptively robust --- it flagged 100\% of adaptive NKE at every $\lambda$ --- but its true held-out FPR was 77\%, i.e.\ it was flagging almost everything. In-sample thresholds manufacture phantom robustness. Under honest calibration (all detectors' held-out FPR $\approx0.04$--$0.05$) that robustness evaporates. First, detector choice still matters against \emph{un-adapted} NKE: a non-parametric cosine-$k$NN detector \citep{sun2022knn} flags \textbf{0\%} of standard NKE (which retains high cosine similarity to training features), so only whitened-distance (Mahalanobis-type) detectors catch NKE at all. Second, evasion generalises across design: penalising the tied-covariance penultimate detector alone already drives the \emph{per-class} (untied) detector on the same features to 0\% as well, and a two-layer ensemble objective drives \textbf{every} parametric detector --- tied and untied, both layers --- to 0\% at 100\% success. \textbf{No feature-distance detector we tested resists an adaptive attacker that can name and differentiate it}; the negative result of Table~\ref{tab:adaptive} is a property of feature-distance detection in general, not of the particular metric. Finally, the evasion is not specific to a non-robust model: repeating the attack against the adversarially trained ResNet-18 (45\% PGD robust accuracy, Sec.~\ref{sec:res-def}) reproduces the no-cost evasion at small $\lambda$ (at $\lambda{=}0.001$: success 1.00, Mahalanobis 100\%$\to$0\% on both white-box and held-out detectors, $L_2\approx12.5$); at larger $\lambda$ success plateaus near 0.53 rather than 0.11, but small-$\lambda$ evasion is unaffected.

\begin{table}[h]
\setlength{\tabcolsep}{2pt} % default is usually 6pt
\centering
\footnotesize
\caption{Detection of adaptive NKE by a battery of feature-distance detectors, each at a \emph{held-out-calibrated} 5\% FPR (CIFAR-10, ResNet-18, $L_2\approx12$; adapt rows at $\lambda{=}0.01$). Row~1 confirms honest calibration (clean FPR $\approx$0.05). Attacking only the tied-penultimate detector also evades the untied one; a two-layer ensemble objective evades every parametric detector. Cosine-$k$NN never flags NKE at all.}
\label{tab:detectors}
\begin{tabular}{l c cccc}
\toprule
Scenario & Success & tied-penult & untied-penult & Maha layer-3 & cosine-$k$NN \\
\midrule
Clean false-positive rate                       & ---  & 0.04 & 0.04 & 0.05 & 0.05 \\
Standard NKE (no adaptation)                    & 1.00 & 1.00 & 0.95 & 1.00 & 0.00 \\
Adapt vs.\ tied-penult                          & 1.00 & \textbf{0.00} & \textbf{0.00} & 0.22 & 0.00 \\
Adapt vs.\ tied ensemble (penult\,+\,layer-3)   & 1.00 & \textbf{0.00} & \textbf{0.00} & \textbf{0.00} & 0.00 \\
\bottomrule
\end{tabular}
\end{table}

\subsection{Extended analysis of \citeauthor{nie2025new}'s variants and transferability (Sec.~\ref{sec:res-ext})}
\textbf{Variants at matched magnitude.} \citeauthor{nie2025new} report success-rate gaps between variants (\eg NMI-FGSM $>$ NI-FGM); but those attacks also differ in the perturbation magnitude they induce. Projecting all four onto the \emph{same} $L_2$ sphere ($L_2\approx 8$ on CIFAR-10), so they differ only in step rule and momentum, the reported differences largely vanish: all four reach $\ge$99.5\% success, all are flagged 100\% by Mahalanobis and 0--3\% by MSP, and edge preservation is nearly identical (Table~\ref{tab:variants}). Only the human-proxy gap varies modestly (sign-based I-FGSM is the most human-destructive at 0.62; $L_2$-normalized NMI-FGM the least at 0.83). The variant mainly trades off perturbation magnitude (its unconstrained $L_2$ ranges $8\to28$ across variants at fixed $\epsilon_l$), and magnitude in turn drives the human gap.

\begin{table}[h]
\setlength{\tabcolsep}{2pt} % default is usually 6pt
\centering
\footnotesize
\caption{Attack variants at matched $L_2\!\approx\!8$ (CIFAR-10, ResNet-18). Success and detectability are invariant to variant; only the human proxy varies.}
\label{tab:variants}
\begin{tabular}{l ccccc}
\toprule
Variant & Success & Maha detect & MSP detect & Human proxy & Edge preservation \\
\midrule
I-FGSM ($L_\infty$ step)   & 1.00 & 1.00 & 0.03 & 0.62 & 0.50 \\
NI-FGM ($L_2$ step)        & 1.00 & 1.00 & 0.00 & 0.78 & 0.49 \\
NMI-FGSM (mom.+sign)       & 1.00 & 1.00 & 0.00 & 0.77 & 0.48 \\
NMI-FGM (mom.+$L_2$)       & 1.00 & 1.00 & 0.00 & 0.83 & 0.46 \\
\bottomrule
\end{tabular}
\end{table}

\textbf{What the transferability collapse reveals.} \citeauthor{nie2025new} found black-box transfer collapses ($<$5\% on ImageNet) but did not explain \emph{which} examples transfer. We test whether source-model manifold-proximity (negative Mahalanobis distance) or edge preservation predicts transfer to a second architecture (ResNet-18$\to$VGG-11). At \emph{small} perturbations, proximity predicts transfer well above chance (AUROC 0.73 at $\epsilon_l=2.5$): the examples that transfer are those still near the manifold. But this predictivity \textbf{decays monotonically to chance ($\approx$0.5 at $\epsilon_l=8$) and reverses at large $\epsilon_l$} (AUROC 0.39 at $\epsilon_l=16$), and edge preservation never predicts transfer ($|r|\le 0.12$). Thus in the large-$\epsilon_l$ regime that defines NKE, transfer becomes disconnected from both manifold-distance and shape.

\begin{figure}[h]
\centering
\includegraphics[width=0.72\linewidth]{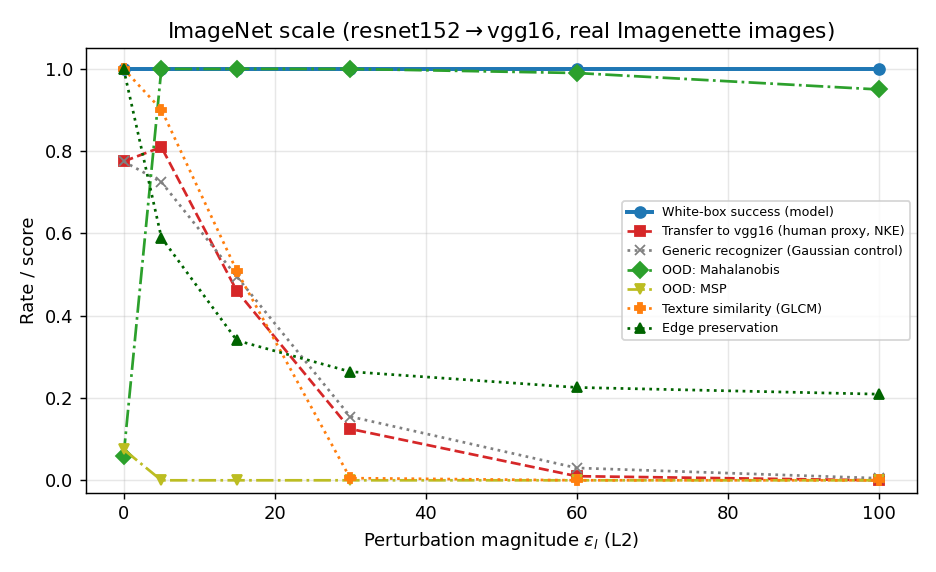}
\caption{ImageNet scale (ResNet-152$\to$VGG-16, real Imagenette images). White-box success/confidence flat at 1.0; transfer (= generic-recognizer proxy) collapses to 0; Mahalanobis detects $\sim$100\% while MSP stays at 0; texture is destroyed while edges partially survive. Note the NKE-transfer (red) and Gaussian-control (gray) curves nearly coincide --- unlike CIFAR-10, the cross-model proxy cannot separate structured from random degradation here (see Sec.~\ref{sec:res-imagenet}).}
\label{fig:imagenet}
\end{figure}

\begin{figure}[h]
\centering
\includegraphics[width=\linewidth]{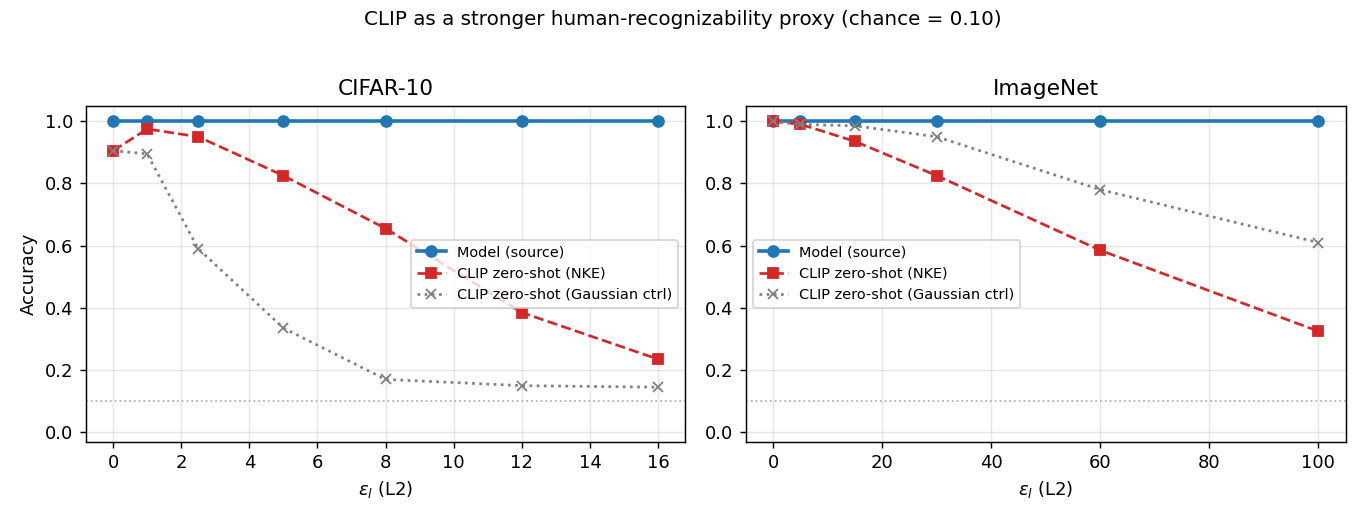}
\caption{CLIP zero-shot accuracy (a non-CNN, broadly-trained recognizability proxy) vs.\ $\epsilon_l$. Model stays at 1.0; CLIP falls far below at both scales (gap confirmed). CIFAR-10: NKE $>$ Gaussian control (not signal loss). ImageNet: NKE $<$ Gaussian control (perturbation is \emph{more} destructive to semantics than noise). Chance $=0.10$.}
\label{fig:clip}
\end{figure}

\section{Vision Transformer shape-bias probe: full results}
\label{app:vit}
This appendix reports the per-track numbers underlying Sec.~\ref{sec:res-vit}. The ViT is a from-scratch model (patch size 4 $\Rightarrow$ 64 tokens, embedding dim 192, depth 6, 3 attention heads, MLP ratio 2; 1.8M parameters), trained for 70 epochs with AdamW (lr $5{\times}10^{-4}$, weight decay 0.05, 5-epoch linear warmup, label smoothing 0.1) under the same crop+flip augmentation as the CNNs, reaching \textbf{83.2\%} clean test accuracy. It is registered as a third CIFAR-10 architecture and passed through the identical pipeline (attack, stimulus set, and every evaluation track) as ResNet-18 and VGG-11. All attacks use NMI-FGM at 60 steps as in the main text; $n{=}200$ correctly-classified, class-stratified images per source.

\begin{figure}[h]
\centering
\includegraphics[width=\linewidth]{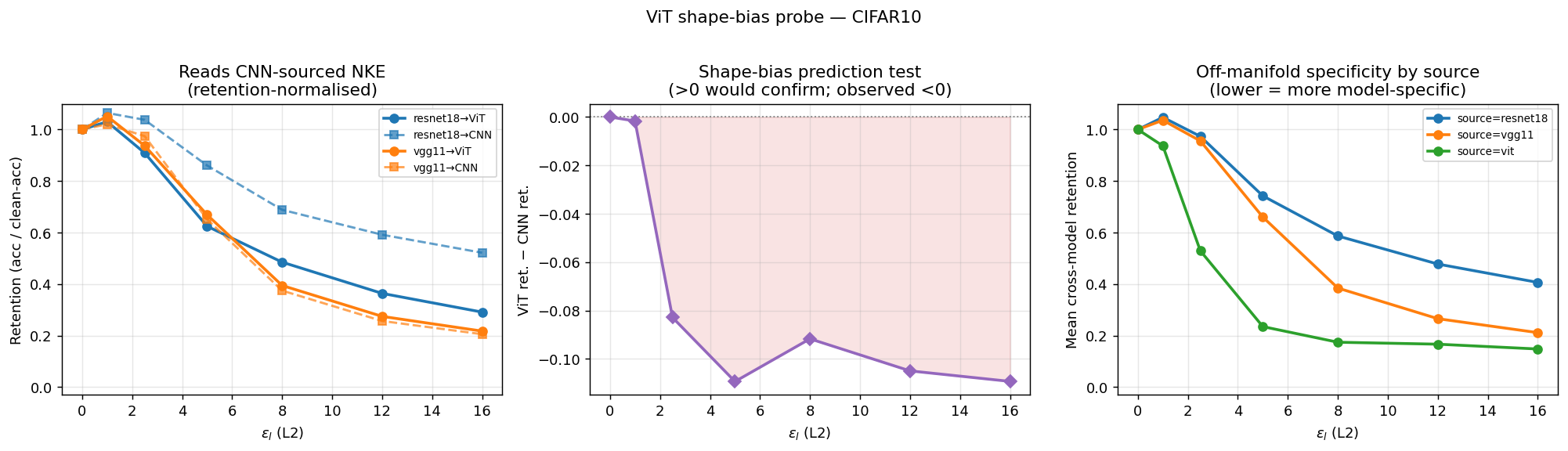}
\caption{ViT shape-bias probe (CIFAR-10), retention-normalised. \textbf{Left:} retention of CNN-sourced NKE by the ViT vs.\ the other CNN recognizer. \textbf{Middle:} the shape-bias prediction test --- ViT-minus-CNN retention is \emph{negative} throughout, so the shape-biased model is \emph{not} a better reader (prediction falsified). \textbf{Right:} off-manifold specificity by source --- NKE generated by the ViT (green) transfers worst and collapses earliest, i.e.\ attention models produce the most model-specific perturbations.}
\label{fig:vit}
\end{figure}

\textbf{Cross-model transferability.} Table~\ref{tab:vit-matrix} extends the black-box matrix (Table~\ref{tab:baseline-matrix}) to all three CIFAR-10 architectures. White-box success is 1.00 on the diagonal; every off-diagonal (black-box) entry is low, and the ViT row (NKE \emph{generated by} the ViT) is the lowest of all.

\begin{table}[h]
\centering
\small
\caption{CIFAR-10 cross-model label-kept matrix at $\epsilon_l{=}12$ (fraction of source-generated NKE the target still labels correctly). Rows: source; columns: target. $*$ = white-box.}
\label{tab:vit-matrix}
\begin{tabular}{l ccc}
\toprule
Source $\downarrow$ / Target $\rightarrow$ & ResNet-18 & VGG-11 & ViT \\
\midrule
ResNet-18 & 1.00* & 0.50 & 0.25 \\
VGG-11    & 0.23  & 1.00* & 0.22 \\
ViT       & 0.13  & 0.16  & 1.00* \\
\bottomrule
\end{tabular}
\end{table}

\textbf{Retention-normalised cross-recognizer analysis.} Table~\ref{tab:vit-retention} gives the retention (accuracy on NKE $\div$ accuracy on that source's clean images, i.e.\ normalised per recognizer to remove the clean-accuracy confound) that underlies Fig.~\ref{fig:vit}. The shape-bias contrast (last two rows) is negative at every $\epsilon_l\ge2.5$: the ViT recognizer is not a better reader of CNN-sourced NKE. The source-specificity block shows ViT-sourced NKE collapsing earliest and lowest.

\begin{table}[h]
\centering
\small
\caption{Retention (acc on NKE / acc on clean) on CIFAR-10. \emph{Top:} shape-bias contrast --- mean (ViT-recognizer retention $-$ other-CNN-recognizer retention) for CNN-sourced NKE; negative means the shape-biased ViT reads CNN NKE \emph{worse}. \emph{Bottom:} source specificity --- mean cross-model retention of the NKE each source generates (lower = more model-specific).}
\label{tab:vit-retention}
\begin{tabular}{l ccccc}
\toprule
$\epsilon_l$ & 2.5 & 5 & 8 & 12 & 16 \\
\midrule
\multicolumn{6}{l}{\emph{Shape-bias prediction test (CNN-sourced NKE)}}\\
ViT $-$ CNN retention (contrast) & $-0.08$ & $-0.11$ & $-0.09$ & $-0.11$ & $-0.11$ \\
\midrule
\multicolumn{6}{l}{\emph{Source specificity (mean cross-model retention by source)}}\\
ResNet-18 source & 0.98 & 0.74 & 0.59 & 0.48 & 0.41 \\
VGG-11 source    & 0.95 & 0.66 & 0.38 & 0.27 & 0.21 \\
\textbf{ViT source} & \textbf{0.53} & \textbf{0.24} & \textbf{0.18} & \textbf{0.17} & \textbf{0.15} \\
\bottomrule
\end{tabular}
\end{table}

\textbf{Mechanism (ViT source).} Table~\ref{tab:vit-mech} is the ViT analogue of Table~\ref{tab:mech}: the same texture-before-shape dissociation, with model accuracy pinned at 1.00.

\begin{table}[h]
\centering
\small
\caption{Shape vs.\ texture for ViT-sourced NKE (CIFAR-10). Model accuracy is 1.00 at every $\epsilon_l$.}
\label{tab:vit-mech}
\begin{tabular}{l ccccc}
\toprule
$\epsilon_l$ & 1 & 2.5 & 5 & 8 & 16 \\
\midrule
Edge preservation (Canny F1) & 0.77 & 0.57 & 0.44 & 0.38 & 0.32 \\
Texture similarity (GLCM)    & 0.91 & 0.63 & 0.07 & 0.00 & 0.00 \\
\bottomrule
\end{tabular}
\end{table}

\textbf{Human proxies (ViT source).} Table~\ref{tab:vit-human} is the ViT analogue of Table~\ref{tab:human}; the generic recognizer is ResNet-18. Note that the NKE curve nearly coincides with the matched-Gaussian control (unlike the ResNet-18 source in Table~\ref{tab:human}), i.e.\ ViT NKE is almost as cross-model-destructive as noise.

\begin{table}[h]
\centering
\small
\caption{Recognizability proxies vs.\ $\epsilon_l$ for ViT-sourced NKE (CIFAR-10; generic recognizer = ResNet-18). Model accuracy is 1.00 at every level.}
\label{tab:vit-human}
\begin{tabular}{l ccccc}
\toprule
$\epsilon_l$ & 0 & 2.5 & 5 & 8 & 16 \\
\midrule
Generic recognizer (NKE)        & 0.95 & 0.41 & 0.14 & 0.16 & 0.12 \\
Generic recognizer (Gauss ctrl) & 0.95 & 0.23 & 0.11 & 0.10 & 0.10 \\
Shape recognizer (NKE)          & 0.69 & 0.39 & 0.18 & 0.10 & 0.12 \\
\bottomrule
\end{tabular}
\end{table}

\textbf{OOD detection (ViT source).} Table~\ref{tab:vit-ood} is the ViT analogue of Table~\ref{tab:ood}. The confidence/energy blindness and ECE$\approx$0 reproduce, but Mahalanobis detection on the ViT's global CLS token is far weaker than on the ResNet-18 penultimate (peaking at 0.50 rather than 1.00).

\begin{table}[h]
\centering
\small
\caption{OOD detection rate at a held-out-calibrated 5\% FPR, and ECE, for ViT-sourced NKE (CIFAR-10). Model accuracy 1.00 throughout.}
\label{tab:vit-ood}
\begin{tabular}{l ccccc}
\toprule
$\epsilon_l$ & 1 & 2.5 & 5 & 8 & 16 \\
\midrule
ECE                            & 0.04 & 0.03 & 0.02 & 0.02 & 0.02 \\
MSP detection                  & 0.00 & 0.00 & 0.00 & 0.00 & 0.00 \\
Energy detection               & 0.00 & 0.00 & 0.00 & 0.00 & 0.00 \\
Mahalanobis detection          & 0.01 & 0.14 & 0.33 & 0.41 & 0.50 \\
\bottomrule
\end{tabular}
\end{table}

\textbf{Adaptive attack (ViT source).} Table~\ref{tab:vit-adaptive} is the ViT analogue of Table~\ref{tab:adaptive}. Oblivious NKE is only 43\% detected here (weaker feature-distance signal), and a tiny penalty ($\lambda{=}0.001$) drives both white-box and held-out detectors to $\approx$0 at full success; success collapses faster than on the CNN as $\lambda$ grows, so the no-cost evasion band is narrower but present.

\begin{table}[h]
\centering
\small
\caption{Adaptive attack against the Mahalanobis detector on the ViT (CIFAR-10, $L_2\approx12$). Detection at a held-out-calibrated 5\% FPR.}
\label{tab:vit-adaptive}
\begin{tabular}{l cccccc}
\toprule
$\lambda$ & 0 & 0.001 & 0.01 & 0.02 & 0.03 & 1.0 \\
\midrule
NKE success                     & 1.00 & 1.00 & 0.93 & 0.66 & 0.35 & 0.17 \\
Perturbation $L_2$              & 12.15 & 12.05 & 12.05 & 12.06 & 12.07 & 12.07 \\
Maha detection (white-box)      & 0.43 & 0.00 & 0.00 & 0.00 & 0.00 & 0.00 \\
Maha detection (held-out)       & 0.42 & 0.01 & 0.00 & 0.00 & 0.00 & 0.00 \\
MSP detection                   & 0.00 & 0.01 & 0.03 & 0.12 & 0.17 & 0.00 \\
\bottomrule
\end{tabular}
\end{table}

\end{document}

%% file: math_commands.tex
\usepackage{amsmath,amsfonts,bm}

\def\eqref#1{equation~\ref{#1}}
\def\1{\bm{1}}

\DeclareMathAlphabet{\mathsfit}{\encodingdefault}{\sfdefault}{m}{sl}
\SetMathAlphabet{\mathsfit}{bold}{\encodingdefault}{\sfdefault}{bx}{n}

%% file: references.bib
@inproceedings{szegedy2014intriguing,
  title={Intriguing properties of neural networks},
  author={Szegedy, Christian and Zaremba, Wojciech and Sutskever, Ilya and Bruna, Joan and Erhan, Dumitru and Goodfellow, Ian and Fergus, Rob},
  booktitle={International Conference on Learning Representations (ICLR)},
  year={2014}
}

@inproceedings{goodfellow2015explaining,
  title={Explaining and harnessing adversarial examples},
  author={Goodfellow, Ian J and Shlens, Jonathon and Szegedy, Christian},
  booktitle={International Conference on Learning Representations (ICLR)},
  year={2015}
}

@article{borji2022new,
  title={A new kind of adversarial example},
  author={Borji, Ali},
  journal={arXiv preprint arXiv:2208.02430},
  year={2022}
}

@article{borji2022addressing,
  title={Addressing the topological defects of disentanglement and adversarial robustness: notes on human vision},
  author={Borji, Ali},
  journal={arXiv preprint arXiv:2208.11580},
  year={2022}
}

@inproceedings{nguyen2015deep,
  title={Deep neural networks are easily fooled: High confidence predictions for unrecognizable images},
  author={Nguyen, Anh and Yosinski, Jason and Clune, Jeff},
  booktitle={IEEE Conference on Computer Vision and Pattern Recognition (CVPR)},
  pages={427--436},
  year={2015}
}

@article{nie2025new,
  title={A New Type of Adversarial Examples},
  author={Nie, Xingyang and Xiao, Guojie and Pan, Su and Wang, Biao and Ge, Huilin and Fang, Tao},
  journal={arXiv preprint arXiv:2510.19347},
  year={2025}
}

@inproceedings{kurakin2018adversarial,
  title={Adversarial examples in the physical world},
  author={Kurakin, Alexey and Goodfellow, Ian and Bengio, Samy},
  booktitle={International Conference on Learning Representations (ICLR) Workshop},
  year={2017}
}

@inproceedings{kurakin2017scale,
  title={Adversarial machine learning at scale},
  author={Kurakin, Alexey and Goodfellow, Ian and Bengio, Samy},
  booktitle={International Conference on Learning Representations (ICLR)},
  year={2017}
}

@inproceedings{tramer2018ensemble,
  title={Ensemble adversarial training: Attacks and defenses},
  author={Tram{\`e}r, Florian and Kurakin, Alexey and Papernot, Nicolas and Goodfellow, Ian and Boneh, Dan and McDaniel, Patrick},
  booktitle={International Conference on Learning Representations (ICLR)},
  year={2018}
}

@inproceedings{carlini2017towards,
  title={Towards evaluating the robustness of neural networks},
  author={Carlini, Nicholas and Wagner, David},
  booktitle={IEEE Symposium on Security and Privacy (S\&P)},
  pages={39--57},
  year={2017}
}

@inproceedings{dong2018boosting,
  title={Boosting adversarial attacks with momentum},
  author={Dong, Yinpeng and Liao, Fangzhou and Pang, Tianyu and Su, Hang and Zhu, Jun and Hu, Xiaolin and Li, Jianguo},
  booktitle={IEEE Conference on Computer Vision and Pattern Recognition (CVPR)},
  pages={9185--9193},
  year={2018}
}

@inproceedings{elsayed2018adversarial,
  title={Adversarial examples that fool both computer vision and time-limited humans},
  author={Elsayed, Gamaleldin and Shankar, Shreya and Cheung, Brian and Papernot, Nicolas and Kurakin, Alexey and Goodfellow, Ian and Sohl-Dickstein, Jascha},
  booktitle={Advances in Neural Information Processing Systems (NeurIPS)},
  year={2018}
}

@inproceedings{geirhos2018generalisation,
  title={Generalisation in humans and deep neural networks},
  author={Geirhos, Robert and Temme, Carlos R and Rauber, Jonas and Sch{\"u}tt, Heiko H and Bethge, Matthias and Wichmann, Felix A},
  booktitle={Advances in Neural Information Processing Systems (NeurIPS)},
  year={2018}
}

@article{zhou2019humans,
  title={Humans can decipher adversarial images},
  author={Zhou, Zhenglong and Firestone, Chaz},
  journal={Nature Communications},
  volume={10},
  number={1},
  pages={1334},
  year={2019}
}

@inproceedings{hendrycks2017baseline,
  title={A baseline for detecting misclassified and out-of-distribution examples in neural networks},
  author={Hendrycks, Dan and Gimpel, Kevin},
  booktitle={International Conference on Learning Representations (ICLR)},
  year={2017}
}

@inproceedings{lee2018simple,
  title={A simple unified framework for detecting out-of-distribution samples and adversarial attacks},
  author={Lee, Kimin and Lee, Kibok and Lee, Honglak and Shin, Jinwoo},
  booktitle={Advances in Neural Information Processing Systems (NeurIPS)},
  year={2018}
}

@inproceedings{liu2020energy,
  title={Energy-based out-of-distribution detection},
  author={Liu, Weitang and Wang, Xiaoyun and Owens, John and Li, Yixuan},
  booktitle={Advances in Neural Information Processing Systems (NeurIPS)},
  year={2020}
}

@inproceedings{ovadia2019can,
  title={Can you trust your model's uncertainty? Evaluating predictive uncertainty under dataset shift},
  author={Ovadia, Yaniv and Fertig, Emily and Ren, Jie and Nado, Zachary and Sculley, D and Nowozin, Sebastian and Dillon, Joshua and Lakshminarayanan, Balaji and Snoek, Jasper},
  booktitle={Advances in Neural Information Processing Systems (NeurIPS)},
  year={2019}
}

@inproceedings{madry2018towards,
  title={Towards deep learning models resistant to adversarial attacks},
  author={Madry, Aleksander and Makelov, Aleksandar and Schmidt, Ludwig and Tsipras, Dimitris and Vladu, Adrian},
  booktitle={International Conference on Learning Representations (ICLR)},
  year={2018}
}

@inproceedings{cohen2019certified,
  title={Certified adversarial robustness via randomized smoothing},
  author={Cohen, Jeremy and Rosenfeld, Elan and Kolter, Zico},
  booktitle={International Conference on Machine Learning (ICML)},
  pages={1310--1320},
  year={2019}
}

@inproceedings{guo2018countering,
  title={Countering adversarial images using input transformations},
  author={Guo, Chuan and Rana, Mayank and Cisse, Moustapha and van der Maaten, Laurens},
  booktitle={International Conference on Learning Representations (ICLR)},
  year={2018}
}

@inproceedings{geirhos2018imagenet,
  title={ImageNet-trained CNNs are biased towards texture; increasing shape bias improves accuracy and robustness},
  author={Geirhos, Robert and Rubisch, Patricia and Michaelis, Claudio and Bethge, Matthias and Wichmann, Felix A and Brendel, Wieland},
  booktitle={International Conference on Learning Representations (ICLR)},
  year={2018}
}

@inproceedings{sun2022knn,
  title={Out-of-distribution detection with deep nearest neighbors},
  author={Sun, Yiyou and Ming, Yifei and Zhu, Xiaojin and Li, Yixuan},
  booktitle={International Conference on Machine Learning (ICML)},
  year={2022}
}

@inproceedings{dosovitskiy2021image,
  title={An Image is Worth 16x16 Words: Transformers for Image Recognition at Scale},
  author={Dosovitskiy, Alexey and Beyer, Lucas and Kolesnikov, Alexander and Weissenborn, Dirk and Zhai, Xiaohua and Unterthiner, Thomas and Dehghani, Mostafa and Minderer, Matthias and Heigold, Georg and Gelly, Sylvain and Uszkoreit, Jakob and Houlsby, Neil},
  booktitle={International Conference on Learning Representations (ICLR)},
  year={2021}
}

@inproceedings{naseer2021intriguing,
  title={Intriguing Properties of Vision Transformers},
  author={Naseer, Muzammal and Ranasinghe, Kanchana and Khan, Salman and Hayat, Munawar and Khan, Fahad Shahbaz and Yang, Ming-Hsuan},
  booktitle={Advances in Neural Information Processing Systems (NeurIPS)},
  year={2021}
}

@inproceedings{tuli2021convolutional,
  title={Are Convolutional Neural Networks or Transformers more like human vision?},
  author={Tuli, Shikhar and Dasgupta, Ishita and Grant, Erin and Griffiths, Thomas L},
  booktitle={Proceedings of the Annual Meeting of the Cognitive Science Society (CogSci)},
  year={2021}
}
